\documentclass[runningheads]{llncs}
\usepackage{graphicx}

\usepackage{tikz}
\usepackage{comment}
\usepackage{amsmath,amssymb} 
\usepackage{color}

\usepackage[pagebackref=true,breaklinks=true,colorlinks,bookmarks=false]{hyperref}

\usepackage{xspace}

\makeatletter
\DeclareRobustCommand\onedot{\futurelet\@let@token\@onedot}
\def\@onedot{\ifx\@let@token.\else.\null\fi\xspace}

\def\eg{\emph{e.g}\onedot} 
\def\ie{\emph{i.e}\onedot} 
 
\def\etc{\emph{etc}\onedot}

\makeatother

\usepackage{multirow,makecell}
\usepackage{floatrow}
\usepackage{subfigure}
\usepackage{booktabs}
\usepackage{pifont}

\usepackage[accsupp]{axessibility}  

\usepackage[width=122mm,left=12mm,paperwidth=146mm,height=193mm,top=12mm,paperheight=217mm]{geometry}

\begin{document}
\pagestyle{headings}
\mainmatter
\def\ECCVSubNumber{1951}  

\title{AiATrack: Attention in Attention for Transformer Visual Tracking} 

\titlerunning{AiATrack: Attention in Attention for Transformer Visual Tracking}
%
\author{Shenyuan Gao\inst{1}, 
Chunluan Zhou\inst{2}, 
Chao Ma\inst{3}, 
Xinggang Wang\inst{1}, 
Junsong Yuan\inst{4}}
\authorrunning{Gao et al.}
%
\institute{$^{1}$~Huazhong University of Science and Technology \quad\quad $^{2}$~Wormpex AI Research \\ $^{3}$~Shanghai Jiao Tong University \quad\quad $^{4}$~State University of New York at Buffalo \\
\email{shenyuangao@gmail.com}, \quad \email{czhou002@e.ntu.edu.sg} \\
\email{chaoma@sjtu.edu.cn}, \quad \email{xgwang@hust.edu.cn}, \quad \email{jsyuan@buffalo.edu}}

\maketitle

\begin{abstract}
Transformer trackers have achieved impressive advancements recently, where the attention mechanism plays an important role. However, the independent correlation computation in the attention mechanism could result in noisy and ambiguous attention weights, which inhibits further performance improvement. To address this issue, we propose an attention in attention (AiA) module, which enhances appropriate correlations and suppresses erroneous ones by seeking consensus among all correlation vectors. Our AiA module can be readily applied to both self-attention blocks and cross-attention blocks to facilitate feature aggregation and information propagation for visual tracking. Moreover, we propose a streamlined Transformer tracking framework, dubbed AiATrack, by introducing efficient feature reuse and target-background embeddings to make full use of temporal references. Experiments show that our tracker achieves state-of-the-art performance on six tracking benchmarks while running at a real-time speed. Code and models are publicly available at \href{https://github.com/Little-Podi/AiATrack}{https://github.com/Little-Podi/AiATrack}.
\end{abstract}

\section{Introduction}
Visual tracking is one of the fundamental tasks in computer vision. It has gained increasing attention because of its wide range of applications \cite{marvasti2021deep,fiaz2019handcrafted}. Given a target with bounding box annotation in the initial frame of a video, the objective of visual tracking is to localize the target in successive frames. Over the past few years, Siamese trackers \cite{bertinetto2016fully,li2018high,li2019siamrpn++,zhang2020ocean}, which regards the visual tracking task as a one-shot matching problem, have gained enormous popularity. Recently, several trackers \cite{wang2021transformer,chen2021transformer,yu2021high,cao2021hift,yan2021learning,xing2022siamese} have explored the application of the Transformer \cite{vaswani2017attention} architecture and achieved promising performance.

The crucial components in a typical Transformer tracking framework \cite{wang2021transformer,chen2021transformer,yu2021high} are the attention blocks. As shown in Fig.~\ref{figure-illustration}, the feature representations of the reference frame and search frame are enhanced via self-attention blocks, and the correlations between them are bridged via cross-attention blocks for target prediction in the search frame. The Transformer attention \cite{vaswani2017attention} takes queries and a set of key-value pairs as input and outputs linear combinations of values with weights determined by the correlations between queries and the corresponding keys. The correlation map is computed by the scaled dot products between queries and keys. However, the correlation of each query-key pair is computed independently, which ignores the correlations of other query-key pairs. This could introduce erroneous correlations due to imperfect feature representations or the existence of distracting image patches in a background clutter scene, resulting in noisy and ambiguous attention weights as visualized in Fig.~\ref{figure-visualization}.

To address the aforementioned issue, we propose a novel attention in attention (AiA) module, which extends the conventional attention \cite{vaswani2017attention} by inserting an inner attention module. The introduced inner attention module is designed to refine the correlations by seeking consensus among all correlation vectors. The motivation of the AiA module is illustrated in Fig.~\ref{figure-illustration}. Usually, if a key has a high correlation with a query, some of its neighboring keys will also have relatively high correlations with that query. Otherwise, the correlation might be noise. Motivated by this, we introduce the inner attention module to utilize these informative cues. Specifically, the inner attention module takes the raw correlations as queries, keys, and values and adjusts them to enhance the appropriate correlations of relevant query-key pairs and suppress the erroneous correlations of irrelevant query-key pairs. We show that the proposed AiA module can be readily inserted into the self-attention blocks to enhance feature aggregation and into the cross-attention block to facilitate information propagation, both of which are very important in a Transformer tracking framework. As a result, the overall tracking performance can be improved.

\definecolor{node1}{RGB}{84, 130, 53}
\definecolor{node2}{RGB}{197, 90, 17}
\definecolor{link1}{RGB}{255, 0, 0}
\definecolor{link2}{RGB}{112, 48, 160}

\begin{figure}[t]
\centering
\includegraphics[width=.98\textwidth]{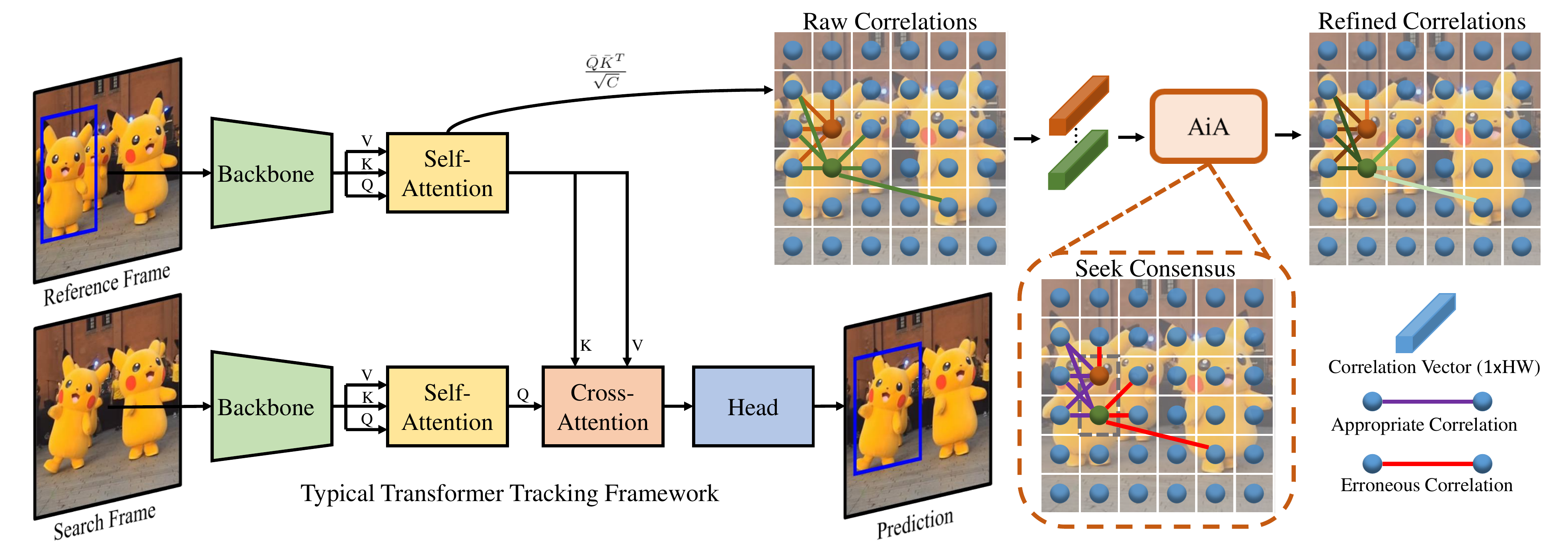}
\caption{Motivation of the proposed method. The left part of the figure shows a typical Transformer tracking framework. On the right, the nodes denote features at different positions in a feature map. These nodes serve as queries and keys for a self-attention block. The links between nodes represent the correlations between queries and keys in the attention mechanism. Some correlations of the \textcolor{node1}{green} node is \textcolor{link1}{erroneous} since it is linked to the nodes at irrelevant positions. By applying the proposed module to the raw correlations, we can seek consensus from the correlations of other nodes (\eg the \textcolor{node2}{brown} node) that can provide supporting cues for the \textcolor{link2}{appropriate} correlations. By this means, the quality of the correlations can be refined.}
\label{figure-illustration}
\end{figure}

How to introduce the long-term and short-term references is still an open problem for visual tracking. With the proposed AiA module, we present AiATrack, a streamlined Transformer framework for visual tracking. Unlike previous practices \cite{zhang2019learning,fu2021stmtrack,wang2021transformer,yan2021learning}, which need an extra computational cost to process the selected reference frame during the model update, we directly reuse the cached features which are encoded before. An IoU prediction head is introduced for selecting high-quality short-term references. Moreover, we introduce learnable target-background embeddings to distinguish the target from the background while preserving the contextual information. With these designs, the proposed AiATrack can efficiently update short-term references and effectively exploit the long-term and short-term references for visual tracking.

We verify the effectiveness of our method by conducting comprehensive experiments on six prevailing benchmarks covering various kinds of tracking scenarios. Without bells and whistles, the proposed AiATrack sets new state-of-the-art results on these benchmarks with a real-time speed of 38 frames per second (fps).

In summary, the main contributions of our work are three-fold:
\begin{itemize}
    \item[$\bullet$] We propose a novel attention in attention (AiA) module, which can mitigate noise and ambiguity in the conventional attention mechanism \cite{vaswani2017attention} and improve tracking performance by a notable margin.
    \item[$\bullet$] We present a neat Transformer tracking framework with the reuse of encoded features and the introduction of target-background embeddings to efficiently and effectively leverage temporal references.
    \item[$\bullet$] We perform extensive experiments and analyses to validate the effectiveness of our designs. The proposed AiATrack achieves state-of-the-art performance on six widely used benchmarks.
\end{itemize}

\section{Related Work}

\subsection{Visual Tracking}
Recently, Transformer \cite{vaswani2017attention} has shown impressive performance in computer vision \cite{carion2020end,zhu2020deformable,dosovitskiy2020image}. It aggregates information from sequential inputs to capture global context by an attention mechanism. Some efforts \cite{yu2020deformable,guo2021graph,fu2021stmtrack} have been made to introduce the attention structure to visual tracking. Recently, several works \cite{wang2021transformer,chen2021transformer,yu2021high,cao2021hift,yan2021learning,xing2022siamese} apply Transformer architecture to visual tracking. Despite their impressive performance, the potential of Transformer trackers is still limited by the conventional attention mechanism. To this end, we propose a novel attention module, namely, attention in attention (AiA), to further unveil the power of Transformer trackers.

How to adapt the model to the appearance change during tracking has also been investigated by previous works \cite{danelljan2019atom,bhat2019learning,zhang2019learning,dai2020high,bhat2020know,fu2021stmtrack,wang2021transformer,yan2021learning}. A straightforward solution is to update the reference features by generation \cite{zhang2019learning} or ensemble \cite{fu2021stmtrack,wang2021transformer,yan2021learning}. However, most of these methods need to resize the reference frame and re-encode the reference features, which may sacrifice computational efficiency. Following discriminative correlation filter (DCF) method \cite{henriques2014high}, another family of approaches \cite{danelljan2019atom,bhat2019learning} optimize the network parameters during the inference. However, they need sophisticated optimization strategies with a sparse update to meet real-time requirements. In contrast, we present a new framework that can efficiently reuse the encoded features. Moreover, a target-background embedding assignment mechanism is also introduced. Different from \cite{ge2021video,yang2021associating,lan2021siamese}, our target-background embeddings are directly introduced to distinguish the target and background regions and provide rich contextual cues.

\subsection{Attention Mechanism}
Represented by non-local operation \cite{wang2018non} and Transformer attention \cite{vaswani2017attention}, attention mechanism has rapidly received great popularity over the past few years. Recently, Transformer attention has been introduced to computer vision as a competitive architecture \cite{carion2020end,zhu2020deformable,dosovitskiy2020image}. In vision tasks, it usually acts as a dynamic information aggregator in spatial and temporal domains. There are some works \cite{huang2019attention,huang2019ccnet} that focus on solving existing issues in the conventional attention mechanism. Unlike these, in this paper, we try to address the noise and ambiguity issue in conventional attention mechanism by seeking consensus among correlations with a global receptive field.

\subsection{Correlation as Feature}
Treating correlations as features has been explored by several previous works \cite{shechtman2007matching,sattler2009scramsac,bian2017gms,rocco2018neighbourhood,li2020correspondence,min2021convolutional,rocco2020efficient,cho2021cats,bhat2020know}. In this paper, we use correlations to refer to the matching results of the pixels or regions. They can be obtained by squared difference, cosine similarity, inner product, \etc. Several efforts have been made to recalibrate the raw correlations by processing them as features through hand-crafted algorithms \cite{sattler2009scramsac,bian2017gms} or learnable blocks \cite{rocco2018neighbourhood,min2021convolutional,li2020correspondence,bhat2020know,rocco2020efficient,cho2021cats}. To our best knowledge, we introduce this insight to the attention mechanism for the first time, making it a unified block for feature aggregation and information propagation in Transformer visual tracking.

\section{Method}

\subsection{Attention in Attention}
\begin{figure}[t]
\centering
\includegraphics[width=.98\textwidth]{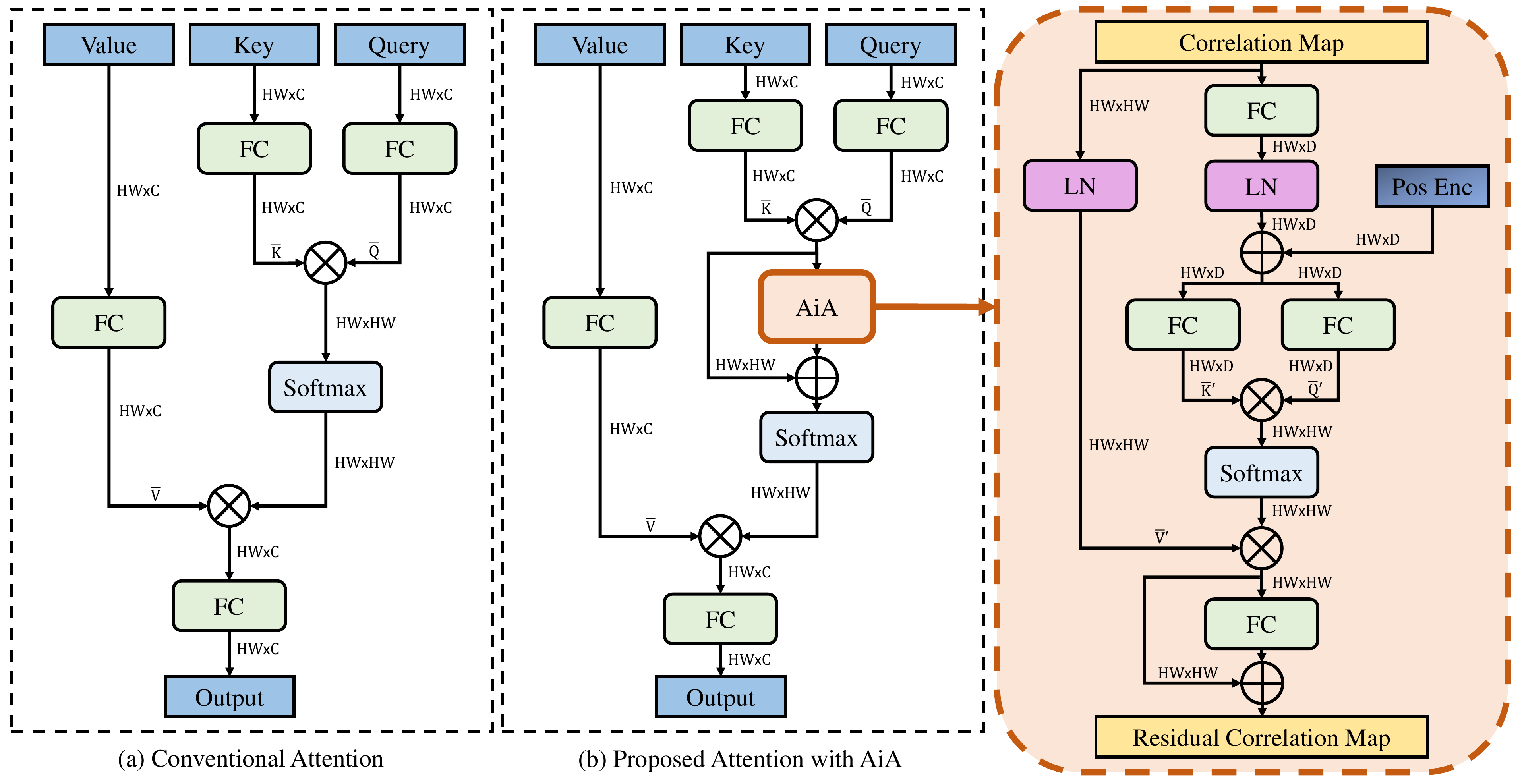}
\caption{Structures of conventional attention and the proposed attention in attention (AiA) module. $\bigotimes$ denotes matrix multiplication and $\bigoplus$ denotes element-wise addition. The numbers beside arrows are feature dimensions which do not include the batch size. Matrix transpose operations are omitted for brevity.}
\label{figure-attention}
\end{figure}

To present our attention in attention module, we first briefly revisit the conventional attention block in vision \cite{dosovitskiy2020image,carion2020end}. As illustrated in Fig.~\ref{figure-attention}(a), it takes a query and a set of key-value pairs as input and produces an output which is a weighted sum of the values. The weights assigned to the values are computed by taking the softmax of the scaled dot products between the query and the corresponding keys. Denote queries, keys and values by $\mathbf{Q}$, $\mathbf{K}$, $\mathbf{V} \in \mathbb{R}^{HW \times C}$ respectively. The conventional attention can be formulated as
\begin{equation}
   \text{ConvenAttn}(\mathbf{Q}, \mathbf{K}, \mathbf{V}) = (\text{Softmax}\left(\frac{\mathbf{\bar{Q}}\mathbf{\bar{K}^{T}}}{\sqrt{C}}\right)\mathbf{\bar{V}})\mathbf{W_{o}}
\end{equation}
where $\mathbf{\bar{Q}} = \mathbf{Q}\mathbf{W_{q}}$, $\mathbf{\bar{K}} = \mathbf{K}\mathbf{W_{k}}$, $\mathbf{\bar{V}} = \mathbf{V}\mathbf{W_{v}}$ are different linear transformations. Here, $\mathbf{W_{q}}$, $\mathbf{W_{k}}$, $\mathbf{W_{v}}$ and $\mathbf{W_{o}}$ denote the linear transform weights for queries, keys, values, and outputs, respectively.

\begin{figure}[t]
\centering
\includegraphics[width=.98\textwidth]{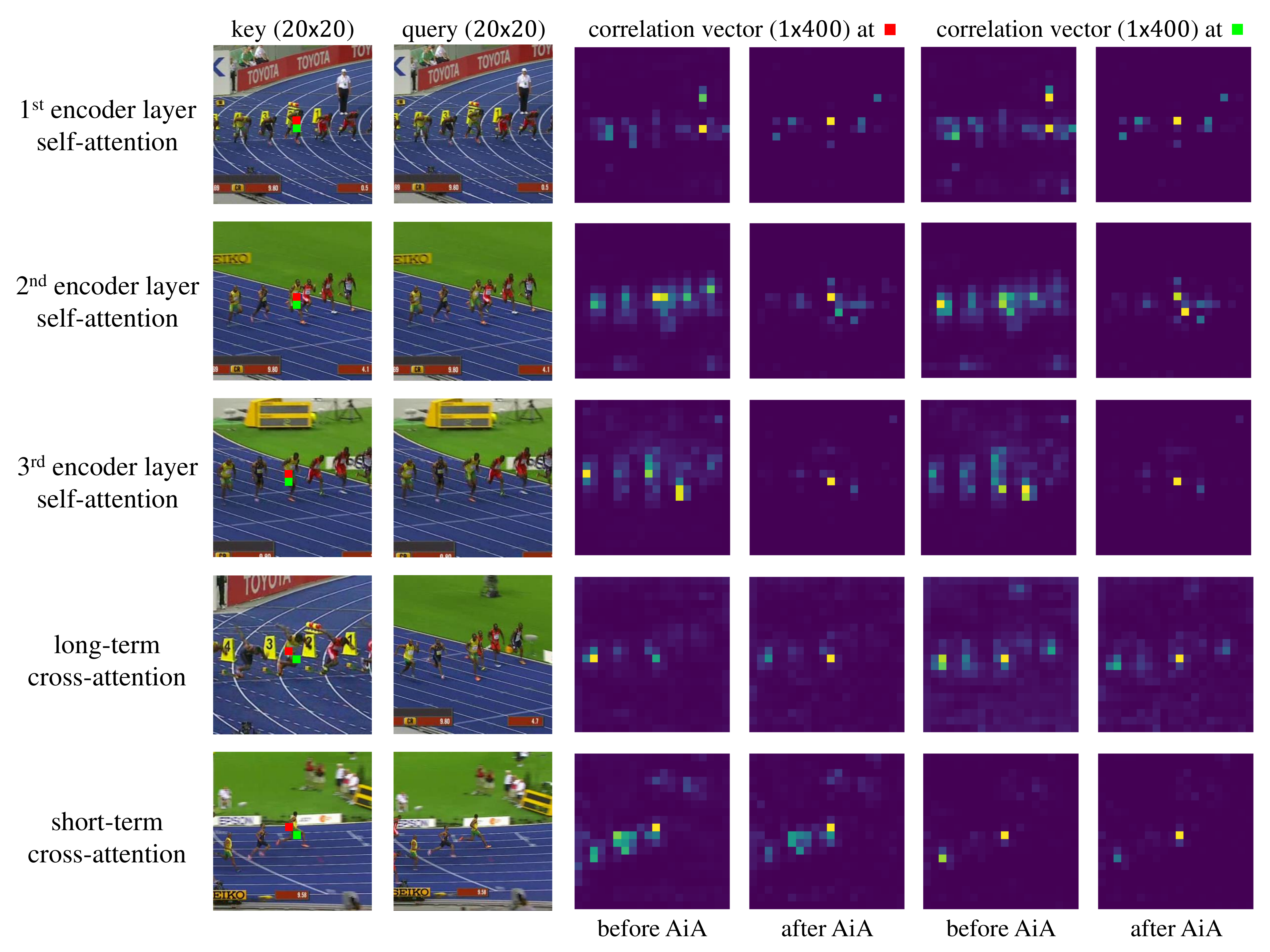}
\caption{Visualization of the effect of the proposed AiA module. We visualize several representative correlation vectors before and after the refinement by the AiA module. The visualized correlation vectors are reshaped according to the spatial positions of queries. We select the correlation vectors of keys corresponding to the target object regions in the first column. It can be observed that the erroneous correlations are effectively suppressed and the appropriate ones are enhanced with the AiA module.}
\label{figure-visualization}
\end{figure}

However, in the conventional attention block, the correlation of each query-key pair in the correlation map $\mathbf{M} = \frac{\mathbf{\bar{Q}}\mathbf{\bar{K}^{T}}}{\sqrt{C}} \in \mathbb{R}^{HW \times HW}$ is computed independently, which ignores the correlations of other query-key pairs. This correlation computation procedure may introduce erroneous correlations due to imperfect feature representations or the existence of distracting image patches in a background clutter scene. These erroneous correlations could result in noisy and ambiguous attentions as visualized in Fig.~\ref{figure-visualization}. They may unfavorably affect the feature aggregation in self-attention and the information propagation in cross-attention, leading to sub-optimal performance for a Transformer tracker.

To address the aforementioned problem, we propose a novel attention in attention (AiA) module to improve the quality of the correlation map $\mathbf{M}$. Usually, if a key has a high correlation with a query, some of its neighboring keys will also have relatively high correlations with that query. Otherwise, the correlation might be a noise. Motivated by this, we introduce the AiA module to utilize the informative cues among the correlations in $\mathbf{M}$. The proposed AiA module seeks the correlation consistency around each key to enhance the appropriate correlations of relevant query-key pairs and suppress the erroneous correlations of irrelevant query-key pairs.

Specifically, we introduce another attention module to refine the correlation map $\mathbf{M}$ before the softmax operation as illustrated in Fig.~\ref{figure-attention}(b). As the newly introduced attention module is inserted into the conventional attention block, we call it an inner attention module, forming an attention in attention structure. The inner attention module itself is a variant of the conventional attention. We consider columns in $\mathbf{M}$ as a sequence of correlation vectors which are taken as queries $\mathbf{Q'}$, keys $\mathbf{K'}$ and values $\mathbf{V'}$ by the inner attention module to output a residual correlation map.

Given the input $\mathbf{Q'}$, $\mathbf{K'}$ and $\mathbf{V'}$, we first generate transformed queries $\mathbf{\bar{Q}'}$ and keys $\mathbf{\bar{K}'}$ as illustrated in the right block of Fig.~\ref{figure-attention}(b). To be specific, a linear transformation is first applied to reduce the dimensions of $\mathbf{Q'}$ and $\mathbf{K'}$ to $HW \times D$ ($D \ll HW$) for computational efficiency. After normalization \cite{ba2016layer}, we add 2-dimensional sinusoidal encoding \cite{dosovitskiy2020image,carion2020end} to provide positional cues. Then, $\mathbf{\bar{Q}'}$ and $\mathbf{\bar{K}'}$ are generated by two different linear transformations. We also normalize $\mathbf{V'}$ to generate the normalized correlation vectors $\mathbf{\bar{V}'}$, \ie $\mathbf{\bar{V}'} = \text{LayerNorm}(\mathbf{V'})$. With $\mathbf{\bar{Q}'}$, $\mathbf{\bar{K}'}$ and $\mathbf{\bar{V}'}$, the inner attention module generates a residual correlation map by
\begin{equation}
    \text{InnerAttn}(\mathbf{M}) = (\text{Softmax}\left(\frac{\mathbf{\bar{Q}'}\mathbf{\bar{K}'^{T}}}{\sqrt{D}}\right)\mathbf{\bar{V}'})(1 + \mathbf{W_{o}'})
\label{equation-aia}
\end{equation}
where $\mathbf{W_{o}'}$ denotes linear transform weights for adjusting the aggregated correlations together with an identical connection.

Essentially, for each correlation vector in the correlation map $\mathbf{M}$, the AiA module generates its residual correlation vector by aggregating the raw correlation vectors. It can be seen as seeking consensus among the correlations with a global receptive field. With the residual correlation map, our attention block with AiA module can be formulated as
\begin{equation}
    \text{AttninAttn}(\mathbf{Q}, \mathbf{K}, \mathbf{V}) = (\text{Softmax}(\mathbf{M} + \text{InnerAttn}(\mathbf{M}))\mathbf{\bar{V}})\mathbf{W_{o}}
\end{equation}

For a multi-head attention block, we share the parameters of the AiA module between the parallel attention heads. It is worth noting that our AiA module can be readily inserted into both self-attenion and cross-attention blocks in a Transformer tracking framework.

\subsection{Proposed Framework}\label{section-network}
\begin{figure}[t]
\centering
\includegraphics[width=.98\textwidth]{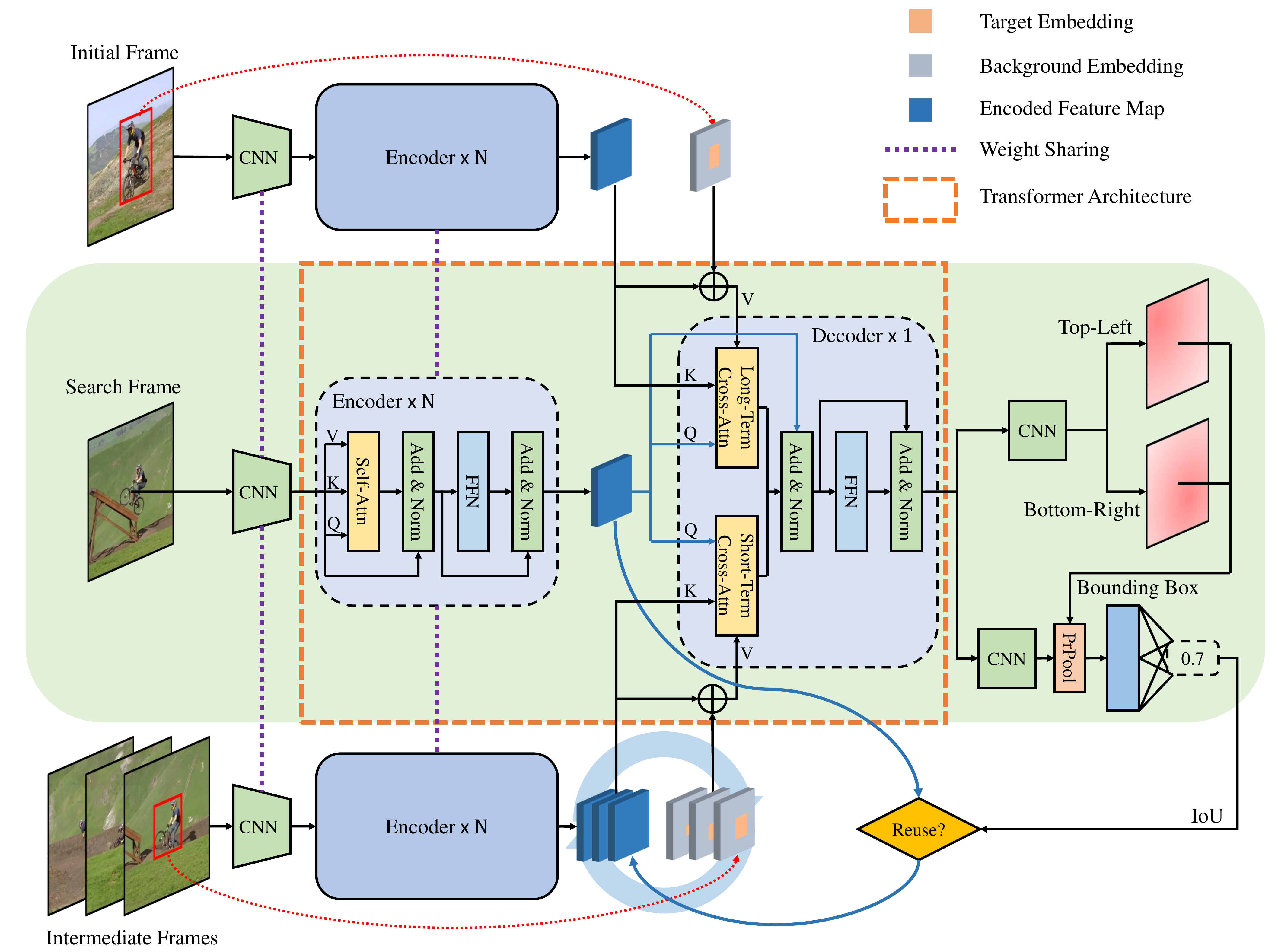}
\caption{Overview of the proposed Transformer tracking framework. The self-attention and cross-attention blocks are all equipped with the proposed AiA module. Note that only the components on the light green background need to be computed during the inference phase as described in Sec.~\ref{section-tracking}.}
\label{figure-framework}
\end{figure}

With the proposed AiA module, we design a simple yet effective Transformer framework for visual tracking, dubbed AiATrack. Our tracker is comprised of a network backbone, a Transformer architecture, and two prediction heads as illustrated in Fig.~\ref{figure-framework}. Given the search frame, the initial frame is taken as a long-term reference and an ensemble of several intermediate frames are taken as short-term references. The features of the long-term and short-term references and the search frame are extracted by the network backbone and then reinforced by the Transformer encoder. We also introduce learnable target-background embeddings to distinguish the target from background regions. The Transformer decoder propagates the reference features as well as the target-background embedding maps to the search frame. The output of the Transformer is then fed to a target prediction head and an IoU prediction head for target localization and short-term reference update, respectively.

\textbf{Transformer Architecture.} The Transformer encoder is adopted to reinforce the features extracted by the convolutional backbone. For the search frame, we flatten its features to obtain a sequence of feature vectors and add sinusoidal positional encoding as in \cite{carion2020end}. The sequence of feature vectors is then taken by the Transformer encoder as its input. The Transformer encoder consists of several layer stacks, each of which is made up of a multi-head self-attention block and a feed-forward network. The self-attention block serves to capture the dependencies among all feature vectors to enhance the original features, and is equipped with the proposed AiA module. Similarly, this procedure is applied independently to the features of the reference frames using the same encoder.

The Transformer decoder propagates the reference information from the long-term and short-term references to the search frame. Different from the classical Transformer decoder \cite{vaswani2017attention}, we remove the self-attention block for simplicity and introduce a two-branch cross-attention design as shown in Fig.~\ref{figure-framework} to retrieve the target-background information from long-term and short-term references. The long-term branch is responsible for retrieving reference information from the initial frame. Since the initial frame has the most reliable annotation of the tracking target, it is crucial for robust visual tracking. However, as the appearance of the target and the background change through the video, the reference information from the long-term branch may not be up-to-date. This could cause tracker drift in some scenes. To address this problem, we introduce the short-term branch to utilize the information from the frames that are closer to the current frame. The cross-attention blocks of the two branches have the identical structure following the query-key-value design in the vanilla transformer \cite{vaswani2017attention}. We take the features of the search frame as queries and the features of the reference frames as keys. The values are generated by combining the reference features with target-background embedding maps, which will be described below. We also insert our AiA module into cross-attention for better reference information propagation.

\textbf{Target-Background Embeddings.} To indicate the target and background regions while preserving the contextual information, we introduce a target embedding $\mathcal{E} ^ {tgt} \in \mathbb{R} ^ {C}$ and a background embedding $\mathcal{E} ^ {bg} \in \mathbb{R} ^ {C}$, both of which are learnable. With $\mathcal{E} ^ {tgt}$ and $\mathcal{E} ^ {bg}$, we generate target-background embedding maps $\mathcal{E} \in \mathbb{R} ^ {HW \times C}$ for the reference frames with a negligible computational cost. Let's consider a location $p$ in a $H \times W$ grid, the embedding assignment is formulated as
\begin{equation}
    \mathcal{E}(p) = \begin{cases} \mathcal{E} ^ {tgt} & \text{if $p$ falls in the target region} \\ \mathcal{E} ^ {bg} & \text{otherwise} \end{cases}
\label{equation-embed}
\end{equation}

Afterward, we attach the target-background embedding maps to the reference features and feed them to cross-attention blocks as values. The target-background embedding maps enrich the reused appearance features by providing contextual cues.

\textbf{Prediction Heads.} As described above, our tracker has two prediction heads. The target prediction head is adopted from \cite{yan2021learning}. Specifically, the decoded features are fed into a two-branch fully-convolutional network which outputs two probability maps for the top-left and the bottom-right corners of the target bounding box. The predicted box coordinates are then obtained by computing the expectations of the probability distributions of the two corners.

To adapt the model to the appearance change during tracking, the tracker needs to keep the short-term references up-to-date by selecting reliable references which contain the target. Moreover, considering our embedding assignment mechanism in Eq.~\ref{equation-embed}, the bounding box of the selected reference frame should be as accurate as possible. Inspired by IoU-Net \cite{jiang2018acquisition} and ATOM \cite{danelljan2019atom}, for each predicted bounding box, we estimate its IoU with the ground truth via an IoU prediction head. The features inside the predicted bounding box are passed to a Precise RoI Pooling layer whose output is taken by a fully connected network to produce an IoU prediction. The predicted IoU is then used to determine whether to include the search frame as a new short-term reference.

We train the two prediction heads jointly. The loss of target prediction is defined by the combination of GIoU loss \cite{rezatofighi2019generalized} and L1 loss between the predicted bounding box and the ground truth. The training examples of the IoU prediction head are generated by sampling bounding boxes around the ground truths. The loss of IoU prediction is defined by mean squared error. We refer readers to the supplementary material for more details about training.

\subsection{Tracking with AiATrack}\label{section-tracking}
Given the initial frame with ground truth annotation, we initialize the tracker by cropping the initial frame as long-term and short-term references and pre-computing their features and target-background embedding maps. For each subsequent frame, we estimate the IoU score of the bounding box predicted by target prediction head for model update. The update procedure is more efficient than the previous practices \cite{fu2021stmtrack,wang2021transformer,yan2021learning}, as we directly reuse the encoded features. Specifically, if the estimated IoU score of the predicted bounding box is higher than the pre-defined threshold, we generate the target-background embedding map for the current search frame and store the embedding map in a memory cache together with its encoded features. For each new-coming frame, we uniformly sample several short-term reference frames and concatenate their features and embedding maps from the memory cache to update the short-term reference ensemble. The latest reference frame in the memory cache is always sampled as it is closest to the current search frame. The oldest reference frame in the memory cache will be popped out if the maximum cache size is reached.

\section{Experiments}

\subsection{Implementation Details}

Our experiments are conducted with NVIDIA GeForce RTX 2080 Ti. We adopt ResNet-50 \cite{he2016deep} as network backbone which is initialized by the parameters pre-trained on ImageNet-1k \cite{deng2009imagenet}. We crop a search patch which is $5 ^ {2}$ times of the target box area from the search frame and resize it to a resolution of $320 \times 320$ pixels. The same cropping procedure is also applied to the reference frames. The cropped patches are then down-sampled by the network backbone with a stride of 16. The Transformer encoder consists of 3 layer stacks and the Transformer decoder consists of only 1 layer. The multi-head attention blocks in our tracker have 4 heads with channel width of 256. The inner AiA module reduces the channel dimension of queries and keys to 64. The FFN blocks have 1024 hidden units. Each branch of the target prediction head is comprised of 5 Conv-BN-ReLU layers. The IoU prediction head consists of 3 Conv-BN-ReLU layers, a PrPool \cite{jiang2018acquisition} layer with pooling size of $3 \times 3$ and 2 fully connected layers.

\begin{table}[t]
\centering
\begin{tabular}{c|c|ccc|ccc|ccc}
\hline
\multirow{2}{*}{Tracker}
& \multirow{2}{*}{Source}
& \multicolumn{3}{c|}{LaSOT \cite{fan2019lasot}}
& \multicolumn{3}{c|}{TrackingNet \cite{muller2018trackingnet}}
& \multicolumn{3}{c}{GOT-10k \cite{huang2019got}} \\
& & AUC & P$_{\text{Norm}}$ & \multicolumn{1}{c|}{P}
& AUC & P$_{\text{Norm}}$ & \multicolumn{1}{c|}{P}
& AO & SR$_{0.75}$ & SR$_{0.5}$ \\
\hline
AiATrack & Ours & 
\textcolor{red}{69.0} & \textcolor{red}{79.4} & \textcolor{red}{73.8} & 
\textcolor{red}{82.7} & \textcolor{red}{87.8} & \textcolor{red}{80.4} & 
\textcolor{red}{69.6} & \textcolor{red}{63.2} & \textcolor{red}{80.0} \\
STARK-ST50 \cite{yan2021learning} & ICCV2021 & 
66.4 & 76.3 & \textcolor{blue}{71.2} & 81.3 & 86.1 & 78.1 & \textcolor{blue}{68.0} & \textcolor{blue}{62.3} & \textcolor{blue}{77.7} \\
KeepTrack \cite{mayer2021learning} & ICCV2021 & 
\textcolor{blue}{67.1} & \textcolor{blue}{77.2} & 70.2 & - & - & - & - & - & - \\
DTT \cite{yu2021high} & ICCV2021 & 
60.1 & - & - & 79.6 & 85.0 & 78.9 & 63.4 & 51.4 & 74.9 \\
TransT \cite{chen2021transformer} & CVPR2021 & 
64.9 & 73.8 & 69.0 & \textcolor{blue}{81.4} & \textcolor{blue}{86.7} & \textcolor{blue}{80.3} & 67.1 & 60.9 & 76.8 \\
TrDiMP \cite{wang2021transformer} & CVPR2021 & 
63.9 & - & 61.4 & 78.4 & 83.3 & 73.1 & 67.1 & 58.3 & \textcolor{blue}{77.7} \\
TrSiam \cite{wang2021transformer} & CVPR2021 & 
62.4 & - & 60.0 & 78.1 & 82.9 & 72.7 & 66.0 & 57.1 & 76.6 \\
KYS \cite{bhat2020know} & ECCV2020 & 
55.4 & 63.3 & - & 74.0 & 80.0 & 68.8 & 63.6 & 51.5 & 75.1 \\
Ocean-online \cite{zhang2020ocean} & ECCV2020 & 
56.0 & 65.1 & 56.6 & - & - & - & 61.1 & 47.3 & 72.1 \\
Ocean-offline \cite{zhang2020ocean} & ECCV2020 & 
52.6 & - & 52.6 & - & - & - & 59.2 & - & 69.5 \\
PrDiMP50 \cite{danelljan2020probabilistic} & CVPR2020 & 
59.8 & 68.8 & 60.8 & 75.8 & 81.6 & 70.4 & 63.4 & 54.3 & 73.8 \\
SiamAttn \cite{yu2020deformable} & CVPR2020 & 
56.0 & 64.8 & - & 75.2 & 81.7 & - & - & - & - \\
DiMP50 \cite{bhat2019learning} & ICCV2019 & 
56.9 & 65.0 & 56.7 & 74.0 & 80.1 & 68.7 & 61.1 & 49.2 & 71.7 \\
SiamRPN++ \cite{li2019siamrpn++} & CVPR2019 & 
49.6 & 56.9 & 49.1 & 73.3 & 80.0 & 69.4 & 51.7 & 32.5 & 61.6 \\
\hline
\end{tabular}
\caption{State-of-the-art comparison on LaSOT, TrackingNet, and GOT-10k. The best two results are shown in \textcolor{red}{red} and \textcolor{blue}{blue}, respectively. All the trackers listed above adopt ResNet-50 pre-trained on ImageNet-1k as network backbone and the results on GOT-10k are obtained without additional training data for fair comparison.}
\label{table-large}
\end{table}

\begin{table}[t]
\centering
\begin{tabular}{c|cccccc}
\hline
\multirow{2}{*}{Tracker}
& SiamRPN++ & PrDiMP50 & TransT & STARK-ST50 & KeepTrack & AiATrack \\
& \cite{li2019siamrpn++} & \cite{danelljan2020probabilistic} & \cite{chen2021transformer} & \cite{yan2021learning} & \cite{mayer2021learning} & (Ours) \\
\hline
NfS30 \cite{kiani2017need} & 50.2 & 63.5 & 65.7 & 65.2 & \textcolor{blue}{66.4} & \textcolor{red}{67.9} \\
OTB100 \cite{wu2015object} & \textcolor{blue}{69.6} & \textcolor{blue}{69.6} & 69.4 & 68.5 & \textcolor{red}{70.9} & \textcolor{blue}{69.6} \\
UAV123 \cite{mueller2016benchmark} & 61.3 & 68.0 & 69.1 & 69.1 & \textcolor{blue}{69.7} & \textcolor{red}{70.6} \\
\hline
Speed (fps) & 35 & 30 & 50 & 42 & 18 & 38 \\
\hline
\end{tabular}
\caption{State-of-the-art comparison on commonly used small-scale datasets in terms of AUC score. The best two results are shown in \textcolor{red}{red} and \textcolor{blue}{blue}.}
\label{table-small}
\end{table}

\subsection{Results and Comparisons}\label{section-comparison}
We compare our tracker with several state-of-the-art trackers on three prevailing large-scale benchmarks (LaSOT \cite{fan2019lasot}, TrackingNet and \cite{muller2018trackingnet} and GOT-10k \cite{huang2019got}) and three commonly used small-scale datasets (NfS30 \cite{kiani2017need}, OTB100 \cite{wu2015object} and UAV123 \cite{mueller2016benchmark}). The results are summarized in Tab.~\ref{table-large} and Tab.~\ref{table-small}.

\begin{figure}[t]
\floatbox[{\capbeside\thisfloatsetup{capbesideposition={left,top},capbesidewidth=.515\textwidth}}]{figure}[\FBwidth]
{\caption{Attribute-based evaluation on LaSOT in terms of AUC score. Our tracker achieves the best performance on all attribute splits while making a significant improvement in various kinds of scenarios such as background clutter, camera motion, and deformation. Axes of each attribute have been normalized.}
\label{figure-attribute}}
{\includegraphics[width=.445\textwidth]{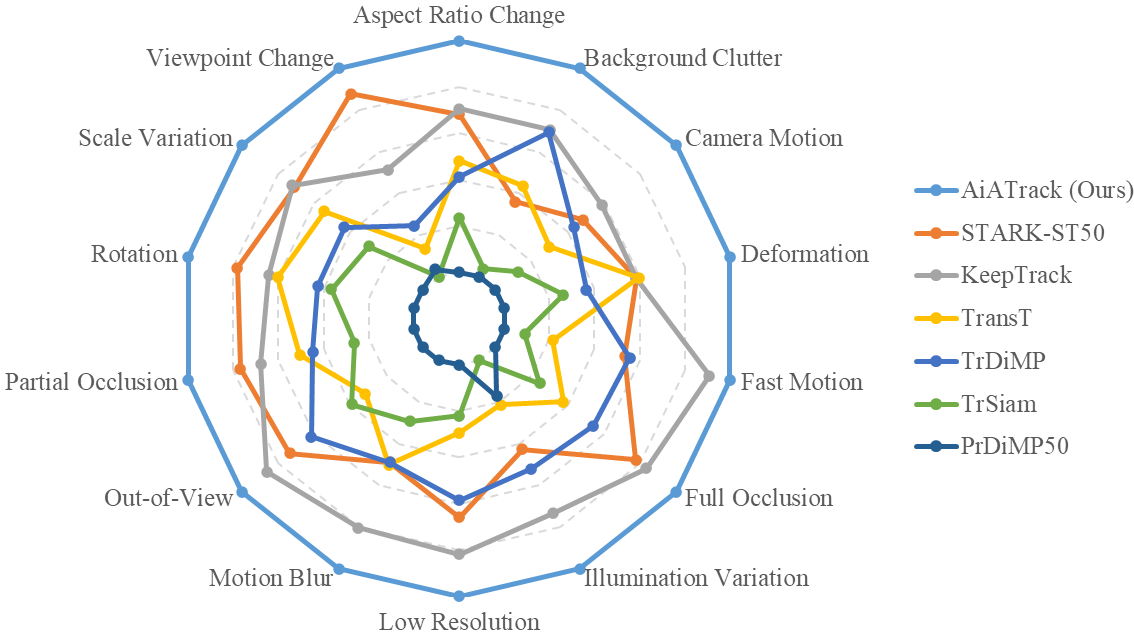}}
\end{figure}

\noindent\textbf{LaSOT.} LaSOT \cite{fan2019lasot} is a densely annotated large-scale dataset, containing 1400 long-term video sequences. As shown in Tab.~\ref{table-large}, our approach outperforms the previous best tracker KeepTrack \cite{mayer2021learning} by 1.9\% in area-under-the-curve (AUC) and 3.6\% in precision while running much faster (see Tab.~\ref{table-small}). We also provide an attribute-based evaluation in Fig.~\ref{figure-attribute} for further analysis. Our method achieves the best performance on all attribute splits. The results demonstrate the promising potential of our approach for long-term visual tracking.

\noindent\textbf{TrackingNet.} TrackingNet \cite{muller2018trackingnet} is a large-scale short-term tracking benchmark. It provides 511 testing video sequences without publicly available ground truths. Our performance reported in Tab.~\ref{table-large} is obtained from the online evaluation server. Our approach achieve 82.7\% in AUC score and 87.8\% in normalized precision score, surpassing all previously published trackers. It demonstrates that our approach is also very competitive for short-term tracking scenarios.

\noindent\textbf{GOT-10k.} To ensure zero overlaps of object classes between training and testing, we follow the one-shot protocol of GOT-10k \cite{huang2019got} and only train our model with the specified subset. The testing ground truths are also withheld and our result is evaluated by the official server. As demonstrated in Tab.~\ref{table-large}, our tracker improves all metrics by a large margin, \eg 2.3\% in success rate compared with STARK \cite{yan2021learning} and TrDiMP \cite{wang2021transformer}, which indicates that our tracker also has a good generalization ability to the objects of unseen classes.

\noindent\textbf{NfS30.} Need for Speed (NfS) \cite{kiani2017need} is a dataset that contains 100 videos with fast-moving objects. We evaluate the proposed tracker on its commonly used version NfS30. As reported in Tab.~\ref{table-small}, our tracker improves the AUC score by 2.7\% over STARK \cite{yan2021learning} and performs the best among the benchmarked trackers.

\noindent\textbf{OTB100.} Object Tracking Benchmark (OTB) \cite{wu2015object} is a pioneering benchmark for evaluating visual tracking algorithms. However, in recent years, it has been noted that this benchmark has become highly saturated \cite{wang2021transformer,yan2021learning,mayer2021learning}. Still, the results in Tab.~\ref{table-small} show that our method can achieve comparable performance with state-of-the-art trackers.

\noindent\textbf{UAV123.} Finally, we report our results on UAV123 \cite{mueller2016benchmark} which includes 123 video sequences captured from a low-altitude unmanned aerial vehicle perspective. As shown in Tab.~\ref{table-small}, our tracker outperforms KeepTrack \cite{mayer2021learning} by 0.9\% and is suitable for UAV tracking scenarios.

\subsection{Ablation Studies}\label{section-ablation}
To validate the importance of the proposed components in our tracker, we conduct ablation studies on LaSOT testing set and its new extension set \cite{fan2021lasot}, totaling 430 diverse videos. We summarize the results in Tab.~\ref{table-ablation}, Tab.~\ref{table-substitute} and Tab.~\ref{table-sensitivity}.

\noindent\textbf{Target-Background Embeddings.} In our tracking framework, the reference frames not only contain features from target regions but also include a large proportion of features from background regions. We implement three variants of our method to demonstrate the necessity of keeping the context and the importance of the proposed target-background embeddings. As shown in the 1st part of Tab.~\ref{table-ablation}, we start from the variant~(a), which is the implementation of the proposed tracking framework with both the target-background embeddings and the AiA module removed. Based on the variant~(a), the variant~(b) further discards the reference features of background regions with a mask. The variant~(c) attaches the target-background embeddings to the reference features. Compared with the variant~(a), the performance of the variant~(b) drops drastically, which suggests that context is helpful for visual tracking. With the proposed target-background embeddings, the variant~(c) can consistently improve the performance over the variant~(a) in all metrics. This is because the proposed target-background embeddings further provide cues for distinguishing the target and background regions while preserving the contextual information.

\noindent\textbf{Long-Term and Short-Term Branch.} As discussed in Sec.~\ref{section-network}, it is important to utilize an independent short-term reference branch to deal with the appearance change during tracking. To validate this, we implement a variant~(d) by removing the short-term branch from the variant~(c). We also implement a variant~(e) by adopting a single cross-attention branch instead of the proposed two-branch design for the variant~(c). Note that we keep the IoU prediction head for these two variants during training to eliminate the possible effect of IoU prediction on feature representation learning. From the 2nd part of Tab.~\ref{table-ablation}, we can observe that the performance of variant~(d) is worse than variant~(c), which suggests the necessity of using short-term references. Meanwhile, compared with variant~(c), the performance of variant~(e) also drops, which validates the necessity to use two separate branches for the long-term and short-term references. This is because the relatively unreliable short-term references may disturb the robust long-term reference and therefore degrade its contribution.

\begin{table}[t]
\centering
\begin{tabular}{ccc|ccc|ccc}
\hline
\multicolumn{3}{c|}{\multirow{2}{*}{Modification}}
& \multicolumn{3}{c|}{LaSOT \cite{fan2019lasot}}
& \multicolumn{3}{c}{LaSOT$_{\text{Ext}}$ \cite{fan2021lasot}} \\
& & & AUC & P$_{\text{Norm}}$ & \multicolumn{1}{c|}{P}
& AUC & P$_{\text{Norm}}$ & P \\
\hline
\multicolumn{1}{c|}{\multirow{3}{*}{1st}} & (a) & none & 
65.8 & 75.8 & 69.5 & 44.5 & 51.5 & 50.5 \\
\multicolumn{1}{c|}{} & (b) & mask & 
64.3 & 72.7 & 66.6 & 42.8 & 50.1 & 48.8 \\
\multicolumn{1}{c|}{} & (c) & embed$^\dagger$ & 
\textbf{67.0} & \textbf{77.0} & \textbf{71.3} & \textbf{44.7} & \textbf{52.7} & \textbf{51.5} \\
\hline
\multicolumn{1}{c|}{\multirow{3}{*}{2nd}} & (d) & w/o short refer & 
66.5 & 76.3 & 70.7 & 44.5 & 51.8 & 50.6 \\
\multicolumn{1}{c|}{} & (e) & w/o branch split & 
63.8 & 72.9 & 66.7 & 42.7 & 50.3 & 48.6 \\
\multicolumn{1}{c|}{} & (c) & w/ both$^\dagger$ & 
\textbf{67.0} & \textbf{77.0} & \textbf{71.3} & \textbf{44.7} & \textbf{52.7} & \textbf{51.5} \\
\hline
\multicolumn{1}{c|}{\multirow{6}{*}{3rd}} & (c) & w/o AiA$^\dagger$ & 
67.0 & 77.0 & 71.3 & 44.7 & 52.7 & 51.5 \\
\multicolumn{1}{c|}{} & (f) & AiA in self-attn & 
68.6 & 78.7 & 72.9 & 46.2 & \textbf{54.4} & 53.4 \\
\multicolumn{1}{c|}{} & (g) & AiA in cross-attn & 
67.5 & 77.9 & 71.8 & 46.2 & 54.2 & 53.3 \\
\multicolumn{1}{c|}{} & (h) & w/o pos in both & 
68.0 & 78.2 & 72.7 & 46.2 & 54.0 & 53.0 \\
\multicolumn{1}{c|}{} & (i) & AiA in both$^\ddagger$ & 
\textbf{68.7} & \textbf{79.3} & \textbf{73.7} & \textbf{46.8} & \textbf{54.4} & \textbf{54.2} \\
\hline
\end{tabular}
\caption{Ablative experiments about different components in the proposed tracker. We use $\dagger$ to denote the basic framework and $\ddagger$ to denote our final model with AiA. The best results in each part of the table are marked in \textbf{bold}.}
\label{table-ablation}
\end{table}

\noindent\textbf{Effectiveness of the AiA Module.} We explore several ways of applying the proposed AiA module to the proposed Transformer tracking framework. The variant~(f) inserts the AiA module into self-attention blocks in the Transformer encoder. Compared with the variant~(c), the performance can be greatly improved on the two subsets of LaSOT. The variant~(g) inserts the AiA module into the cross-attention blocks in the Transformer decoder, which also brings a consistent improvement. These two variants demonstrate that the AiA module generalizes well to both self-attention blocks and cross-attention blocks. When we apply the AiA module to both self-attention blocks and cross-attention blocks, \ie the final model (i), the performance on the two subsets of LaSOT can be improved by 1.7$\sim$2.7\% in all metrics compared with the basic framework (c).

Recall that we introduce positional encoding to the proposed AiA module (see Fig.~\ref{figure-attention}). To verify its importance, we implement a variant~(h) by removing positional encoding from the variant~(i). We can observe that the performance drops accordingly. This validates the necessity of positional encoding, as it provides spatial cues for consensus seeking in the AiA module. More analysis about the components of the AiA module are provided in the supplementary material.

\noindent\textbf{Superiority of the AiA Module.} One may concern that the performance gain of the AiA module is brought by purely adding extra parameters. Thus, we design two other variants to demonstrate the superiority of the proposed module.

First, we implement a variant of our basic framework where each Attention-Add-Norm block is replaced by two cascaded ones. From the comparison of the first two rows in Tab.~\ref{table-substitute}, we can observe that simply increasing the number of attention blocks in our tracking framework does not help much, which demonstrates that our AiA module can further unveil the potential of the tracker.

We also implement a variant of our final model by replacing the proposed inner attention with a convolutional bottleneck \cite{he2016deep}, which is designed to have a similar computational cost. From the comparison of the last two rows in Tab.~\ref{table-substitute}, we can observe that inserting a convolutional bottleneck can also bring positive effects, which suggests the necessity of correlation refinement. However, the convolutional bottleneck can only perform a fixed aggregation in each local neighborhood, while our AiA module has a global receptive field with dynamic weights determined by the interaction among correlation vectors. As a result, our AiA module can seek consensus more flexibly and further boost the performance.

\begin{table}[t]
\centering
\begin{tabular}{c|c|ccc|ccc|c}
\hline
\multirow{2}{*}{Modification}
& Correlation
& \multicolumn{3}{c|}{LaSOT \cite{fan2019lasot}}
& \multicolumn{3}{c|}{LaSOT$_{\text{Ext}}$ \cite{fan2021lasot}}
& Speed \\
& Refinement
& AUC & P$_{\text{Norm}}$ & \multicolumn{1}{c|}{P}
& AUC & P$_{\text{Norm}}$ & P
& (fps) \\
\hline
w/o AiA$^\dagger$ & \multirow{2}{*}{\ding{55}} &
67.0 & 77.0 & 71.3 & 44.7 & 52.7 & 51.5 & 44 \\
w/o AiA cascade & &
67.1 & 77.0 & 71.7 & 44.6 & 52.9 & 51.6 & 40 \\
\hline
conv in both & \multirow{2}{*}{\ding{51}} &
67.9 & 78.2 & 72.8 & 46.0 & 53.4 & 52.8 & 39 \\
AiA in both$^\ddagger$ & &
\textbf{68.7} & \textbf{79.3} & \textbf{73.7} & \textbf{46.8} & \textbf{54.4} & \textbf{54.2} & 38 \\
\hline
\end{tabular}
\caption{Superiority comparison with the tracking performance and the running speed.}
\label{table-substitute}
\end{table}

\begin{table}[t]
\centering
\begin{tabular}{c|ccccccc}
\hline
Ensemble Size & 1 & 2 & 3 & 4 & 5 & 6 & 10 \\
\hline
LaSOT \cite{fan2019lasot} & 66.8 & 68.1 & \colorbox{lightgray}{68.7} & \textbf{69.0} & 68.2 & 68.6 & 68.9 \\
LaSOT$_{\text{Ext}}$ \cite{fan2021lasot} & 44.9 & 46.3 & \colorbox{lightgray}{46.8} & 46.2 & 47.4 & \textbf{47.7} & 47.1 \\
\hline
Speed (fps) & 39 & 39 & 38 & 38 & 38 & 38 & 34 \\
\hline
\end{tabular}
\caption{Impact of ensemble size in terms of AUC score and the running speed. All of our ablative experiments are conducted with ensemble size as 3 by default.}
\label{table-sensitivity}
\end{table}

\noindent\textbf{Visualization Perspective.} In Fig.~\ref{figure-visualization}, we visualize correlation maps from the perspective of keys. This is because we consider the correlations of one key with queries as a correlation vector. Thus, the AiA module performs refinement by seeking consensus among the correlation vectors of keys. Actually, refining the correlations from the perspective of queries also works well, achieving 68.5\% in AUC score on LaSOT.

\noindent\textbf{Short-Term Reference Ensemble.} We also study the impact of the ensemble size in the short-term branch. Tab.~\ref{table-sensitivity} shows that by increasing the ensemble size from 1 to 3, the performance can be stably improved. Further increasing the ensemble size does not help much and has little impact on the running speed.

\section{Conclusion}
In this paper, we present an attention in attention (AiA) module to improve the attention mechanism for Transformer visual tracking. The proposed AiA module can effectively enhance appropriate correlations and suppress erroneous ones by seeking consensus among all correlation vectors. Moreover, we present a streamlined Transformer tracking framework, dubbed AiATrack, by introducing efficient feature reuse and embedding assignment mechanisms to fully utilize temporal references. Extensive experiments demonstrate the superiority of the proposed method. We believe that the proposed AiA module could also be beneficial in other related tasks where the Transformer architecture can be applied to perform feature aggregation and information propagation, such as video object segmentation \cite{yang2021associating,lan2021siamese,duke2021sstvos,mao2021joint}, video object detection \cite{he2021end} and multi-object tracking \cite{sun2020transtrack,meinhardt2021trackformer,zhang2021bytetrack}. \\

\noindent\textbf{Acknowledgment.} This work is supported in part by National Key R\&D Program of China No. 2021YFC3340802, National Science Foundation Grant CNS1951952 and National Natural Science Foundation of China Grant 61906119.

%
%
\bibliographystyle{splncs04}
\bibliography{egbib}

\clearpage

\begin{center}
\textbf{\large AiATrack: Attention in Attention for Transformer Visual Tracking (Supplementary Material)}
\end{center}

The supplementary material provides additional details about the experiments and analyses of the proposed method.

\section{Additional Experiment Details}

\subsection{Target Prediction}
To make the tracking procedure in an end-to-end manner without tedious post-processing, we adopt the anchor-free prediction head proposed in \cite{yan2021learning}, which outputs the probability maps $P_{tl}(x,y)$ and $P_{br}(x,y)$ for the top-left and the bottom-right bounding box corners. The coordinates $\widehat{x}_{tl}$, $\widehat{y}_{tl}$, $\widehat{x}_{br}$, $\widehat{y}_{br}$ of the predicted bounding box are then obtained by 
\begin{gather}
    \widehat{x}_{tl} = \sum_{y = 0}^{H}\sum_{x = 0}^{W}x \cdot P_{tl}(x,y) \text{, } \widehat{y}_{tl} = \sum_{y = 0}^{H}\sum_{x = 0}^{W}y \cdot P_{tl}(x,y) \\
    \widehat{x}_{br} = \sum_{y = 0}^{H}\sum_{x = 0}^{W}x \cdot P_{br}(x,y) \text{, } \widehat{y}_{br} = \sum_{y = 0}^{H}\sum_{x = 0}^{W}y \cdot P_{br}(x,y)
\end{gather}

\subsection{Training Objective}
With the predicted bounding box $\widehat{b}$ and predicted IoU $\widehat{i}$, the whole network is jointly trained by minimizing prediction errors. The bounding box prediction loss is defined as the combination of GIoU loss \cite{rezatofighi2019generalized} and L1 loss. Together with the IoU prediction loss, the loss function can be written as 
\begin{equation}
    L = \lambda_{giou}L_{giou}(b, \widehat{b}) + \lambda_{l1}\Vert b - \widehat{b}\Vert_{1} + \lambda_{mse}(i - \widehat{i})^{2}
\end{equation}
where $b$ and $i$ represent the ground truths of bounding box and IoU respectively and $\lambda_{giou}$, $\lambda_{l1}$, $\lambda_{mse}$ are the trade-off weights.

\subsection{Training Strategy}
Similar to previous works \cite{danelljan2019atom,bhat2019learning,chen2021transformer,wang2021transformer,yan2021learning}, we utilize the training splits of LaSOT \cite{fan2019lasot}, TrackingNet \cite{muller2018trackingnet}, GOT-10k \cite{huang2019got}, and COCO \cite{lin2014microsoft} for offline training. As for the COCO image dataset, we apply data augmentation to generate synthetic video clips of diverse classes. During training, we randomly sample the search frame and reference frames such that the index of the search frame is larger than the indexes of reference frames. For training efficiency, we only sample one frame as the short-term reference. We also apply random affine transformations to jitter the sizes and locations of the short-term reference frame and search frame to simulate real tracking scenarios and avoid the influence of absolute position bias caused by padding \cite{islam2020much,li2019siamrpn++,zhang2019deeper}. The network is trained with the AdamW optimizer \cite{loshchilov2017decoupled}. The learning rate is 1e-5 for the network backbone and 1e-4 for the other components. It decays by a factor of 10 during training. The parameters of the first convolutional layer and the first stage in the ResNet-50 \cite{he2016deep} backbone are fixed during training.

\begin{figure}[t]
\centering
\includegraphics[width=.98\textwidth]{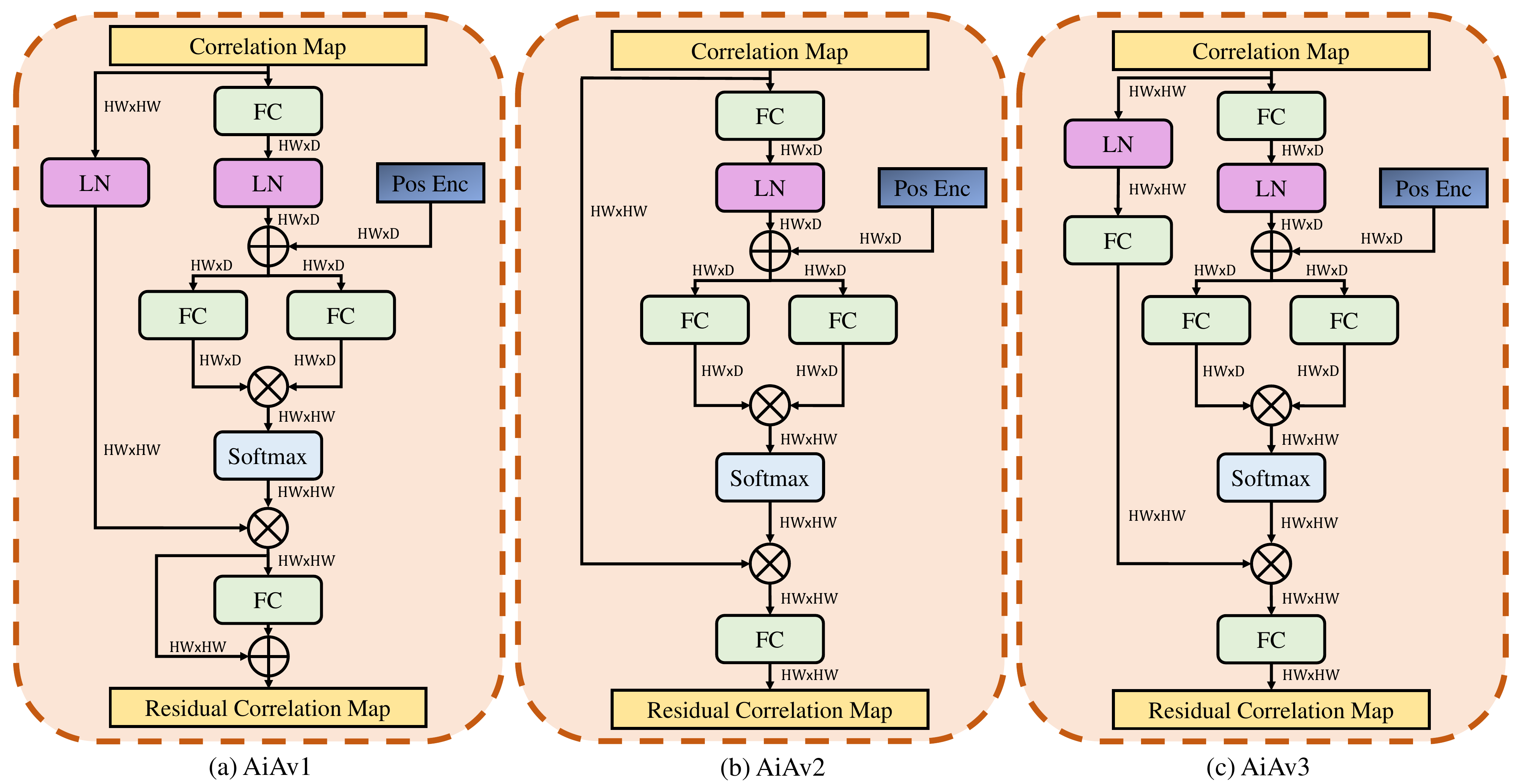}
\caption{Detailed illustration of the differences between different structures of the AiA module. $\bigotimes$ denotes matrix multiplication and $\bigoplus$ denotes element-wise addition. The numbers beside arrows are feature dimensions which do not include the batch size. Matrix transpose operations are omitted for brevity.}
\label{figure-structure}
\end{figure}

\begin{table}[t]
\centering
\begin{tabular}{c|ccc|ccc|ccc}
\hline
\multirow{2}{*}{Structure}
& \multicolumn{3}{c|}{Modification}
& \multicolumn{3}{c|}{LaSOT \cite{fan2019lasot}}
& \multicolumn{3}{c}{LaSOT$_{\text{Ext}}$ \cite{fan2021lasot}} \\
& LN & LT & IC
& AUC & P$_{\text{Norm}}$ & \multicolumn{1}{c|}{P}
& AUC & P$_{\text{Norm}}$ & P \\
\hline
AiAv1 & \ding{51} & & \ding{51} & 68.7 & 79.3 & 73.7 & 46.8 & 54.4 & 54.2 \\
AiAv2 & & & & 68.8 & 79.3 & 73.6 & 46.7 & 54.5 & 53.8 \\
AiAv3 & \ding{51} & \ding{51} & & \textbf{69.2} & \textbf{79.6} & \textbf{74.3} & \textbf{48.4} & \textbf{56.6} & \textbf{56.2} \\
\hline
\end{tabular}
\caption{Study about the different structures of the AiA module. \textbf{LN} denotes applying layer normalization to the value. \textbf{LT} denotes applying linear transformation to the value. \textbf{IC} denotes using identical connection after the correlation aggregation.}
\label{table-structure}
\end{table}

\subsection{Different Structures of the AiA Module}
Besides variant~(h) in the paper, we also explore other structures of the AiA module, where the following components are studied: (1) Layer normalization applied to the value. (2) Linear transformation applied to the value. (3) Identical connection after the correlation aggregation. To evaluate their effect, we design two other structures of the AiA module, \ie AiAv2 and AiAv3. The differences between these structures are shown in Fig.~\ref{figure-structure}. Note that AiAv1 is the structure we implement in AiATrack and AiAv3 is a typical self-attention structure in the vanilla Transformer \cite{vaswani2017attention}.

From the results in Tab.~\ref{table-structure}, we can observe that the layer normalization and the identical connection are not key components in our AiA module. Applying linear transformation to the value can further improve the performance, but we remove it for the trade-off between performance and computational cost. Besides the observations above, all the experimental results validate the effectiveness of correlation refinement in the conventional attention mechanism with an extra attention module.

\subsection{Results on VOT}
Different from previous reset-based evaluation protocol \cite{kristan2019seventh}, VOT2020 \cite{kristan2020eighth} proposes a new anchor-based evaluation protocol which is more reasonable. The same as STARK \cite{yan2021learning} and DualTFR \cite{xie2021learning}, we use Alpha-Refine \cite{yan2021alpha} to generate masks for evaluation since the ground truths of VOT2020 are annotated by the segmentation masks. The overall performance is ranked by the Expected Average Overlap (EAO). As shown in Tab.~\ref{table-vot}, our tracker exhibits very competitive performance, outperforming STARK with a margin of 5\% in terms of EAO.

\begin{table}[t]
\centering
\begin{tabular}{c|ccccc}
\hline
\multirow{2}{*}{Tracker}
& Alpha-Refine & OceanPlus & DualTFR & STARK-ST50 & AiATrack \\
& \cite{yan2021alpha} & \cite{zhang2021toward} & \cite{xie2021learning} & \cite{yan2021learning} & (Ours) \\
\hline
EAO & 0.482 & 0.491 & 0.528 & 0.505 & \textbf{0.530} \\
Accuracy & 0.754 & 0.685 & 0.755 & 0.759 & \textbf{0.764} \\
Robustness & 0.777 & \textbf{0.842} & 0.836 & 0.817 & 0.827 \\
\hline
\end{tabular}
\caption{State-of-the-art comparison on VOT2020.}
\label{table-vot}
\end{table}

\section{Additional Visualization Results}

\begin{figure}[t]
\centering
\subfigure{\includegraphics[width=.28\textwidth]{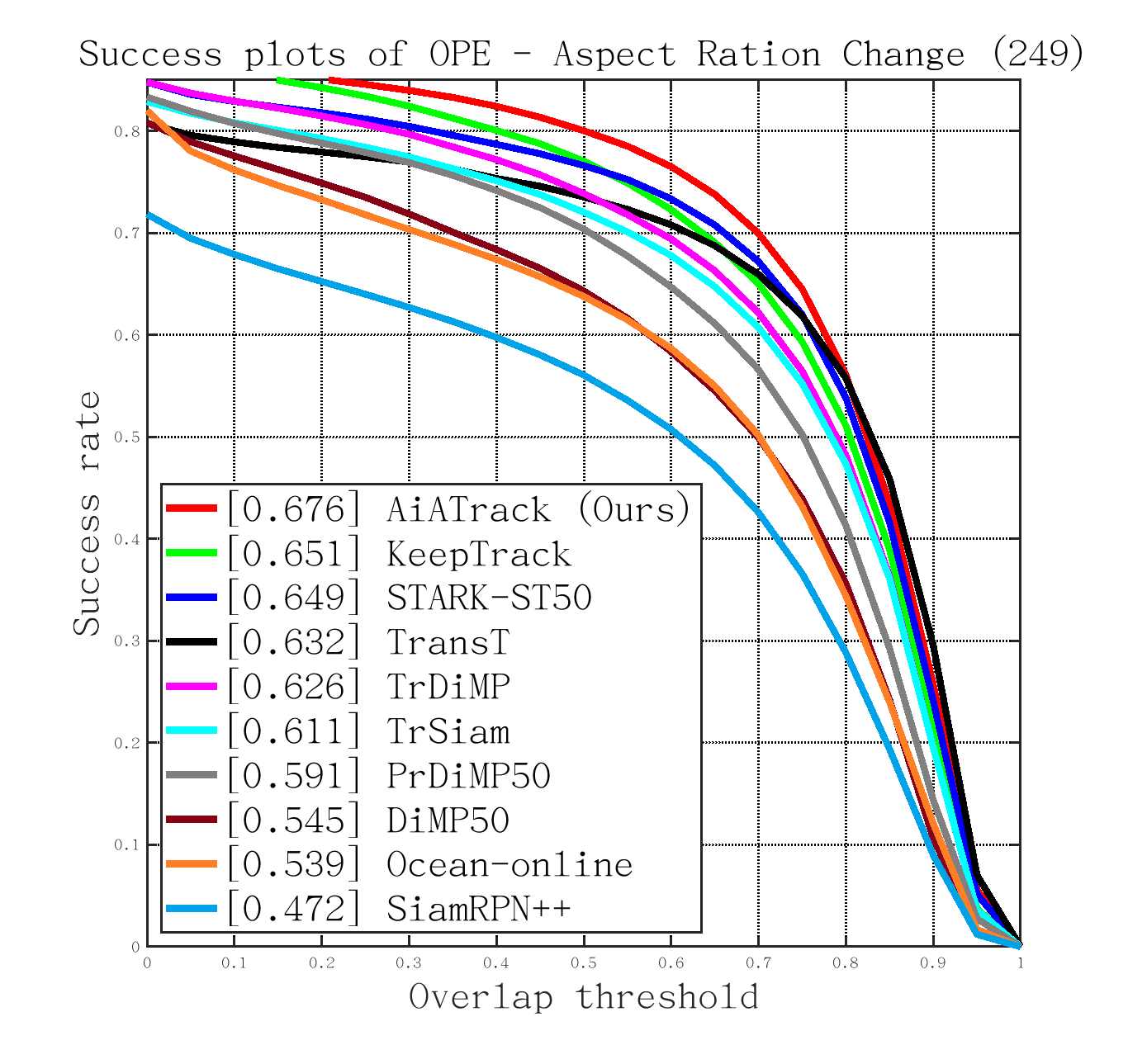}}
\enspace
\subfigure{\includegraphics[width=.28\textwidth]{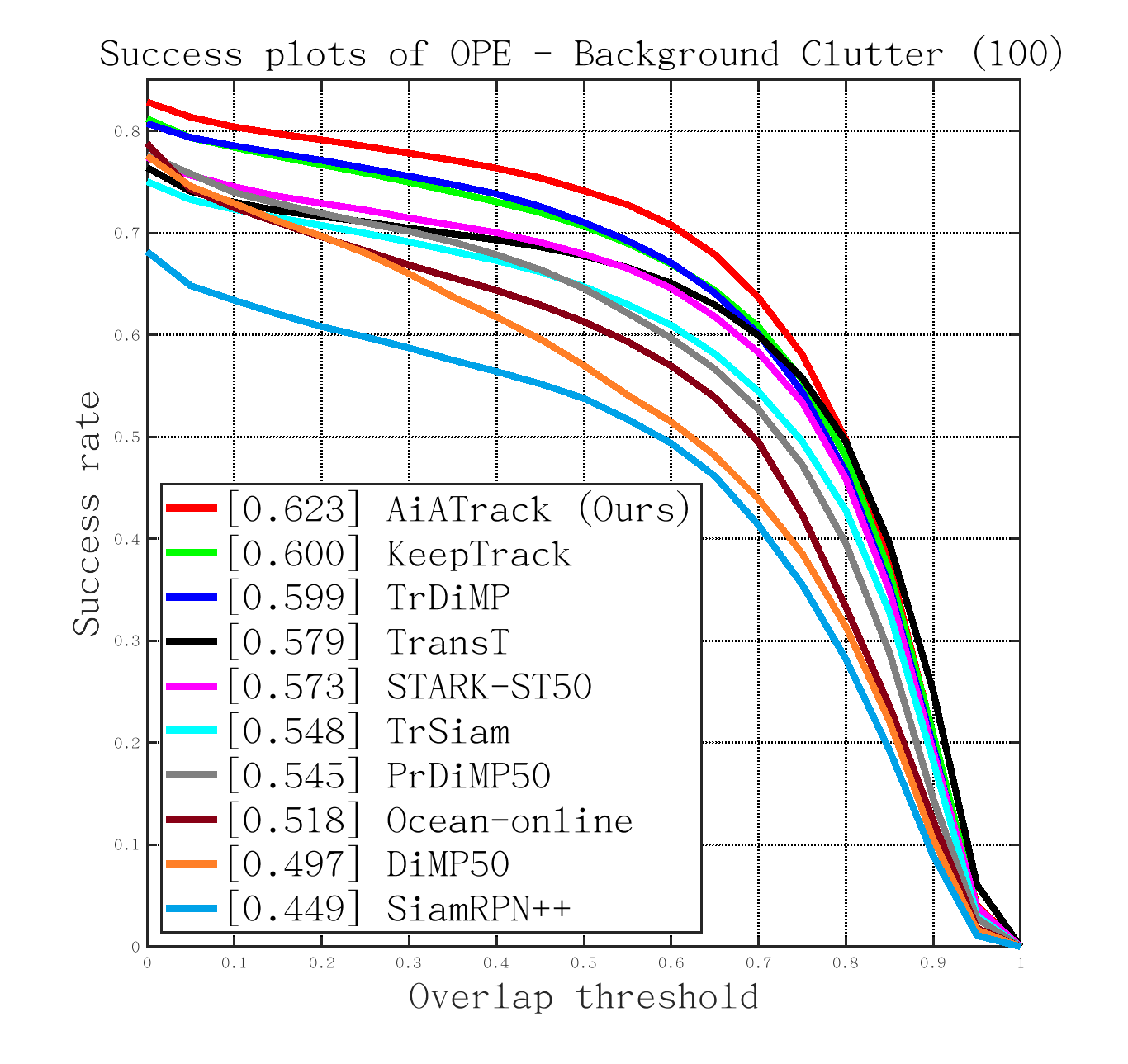}}
\enspace
\subfigure{\includegraphics[width=.28\textwidth]{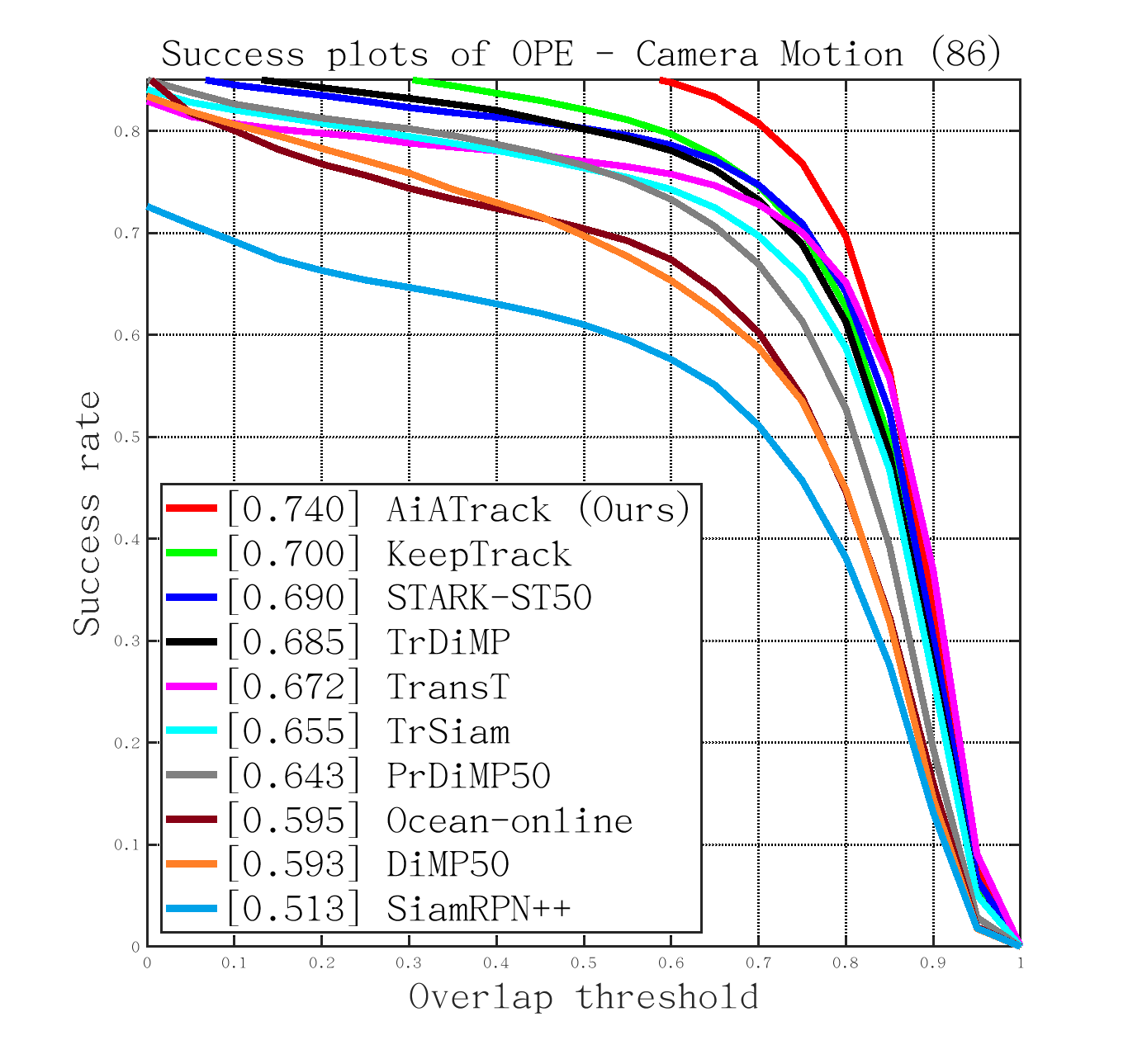}}
\enspace
\subfigure{\includegraphics[width=.28\textwidth]{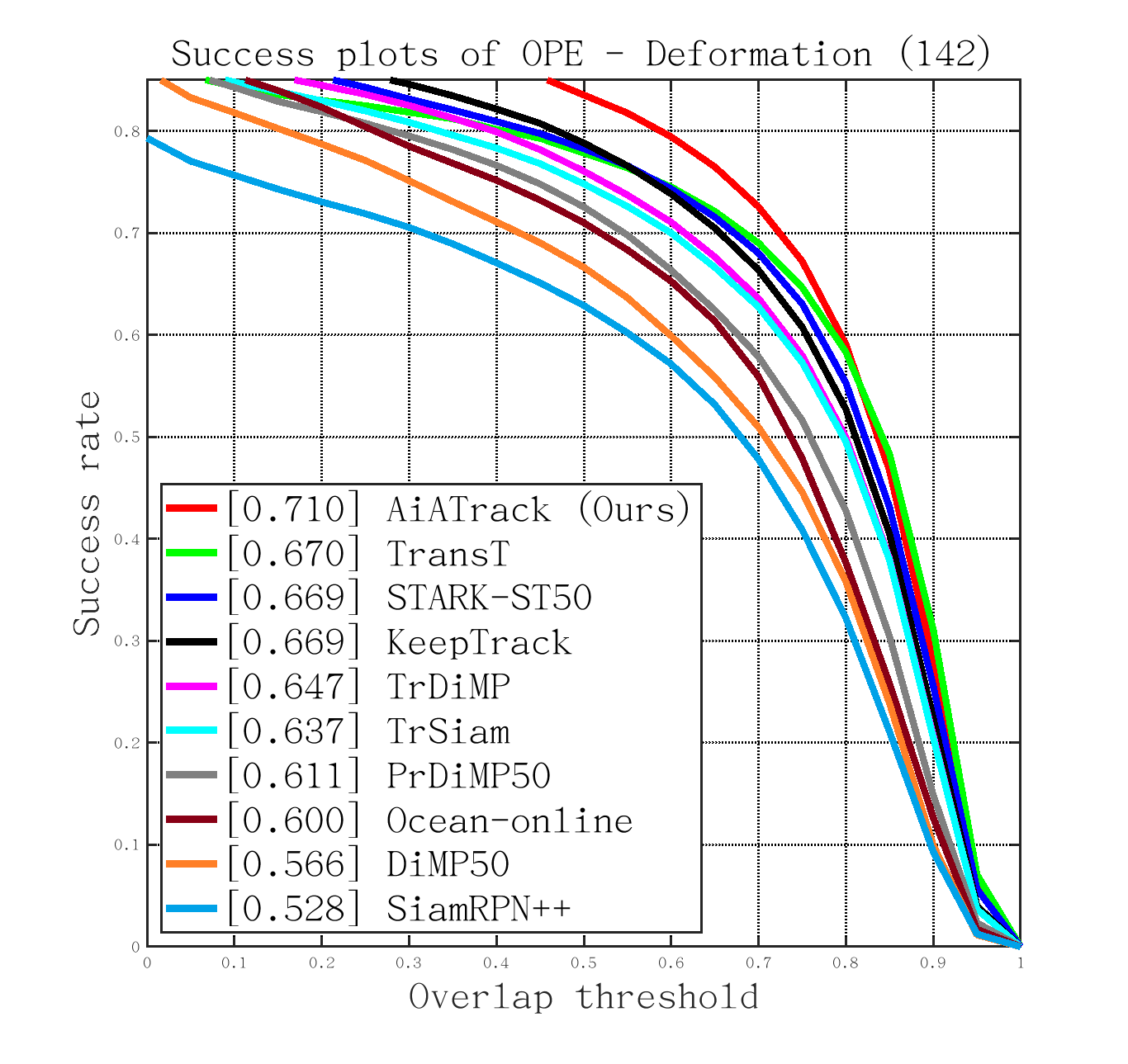}}
\enspace
\subfigure{\includegraphics[width=.28\textwidth]{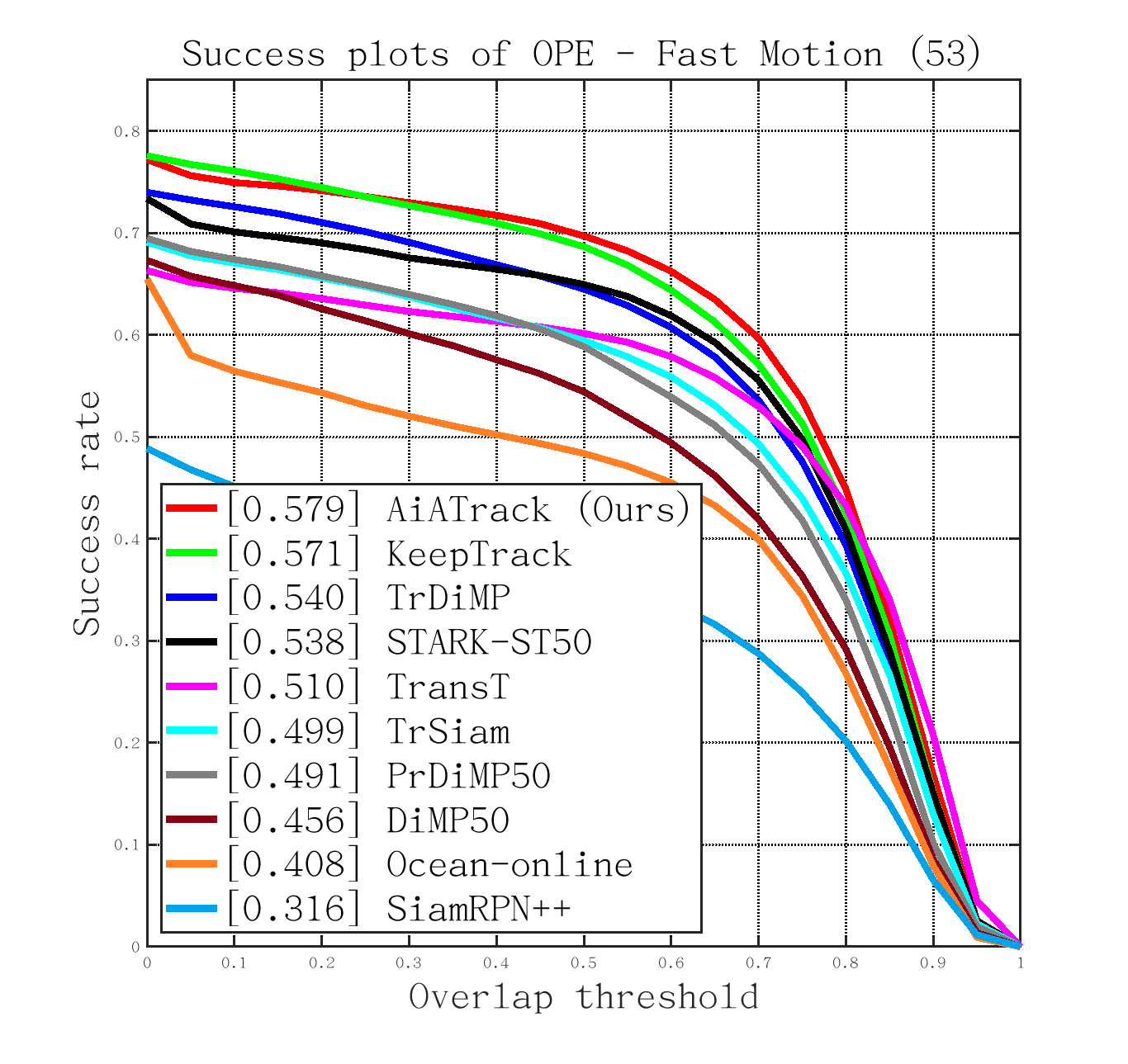}}
\enspace
\subfigure{\includegraphics[width=.28\textwidth]{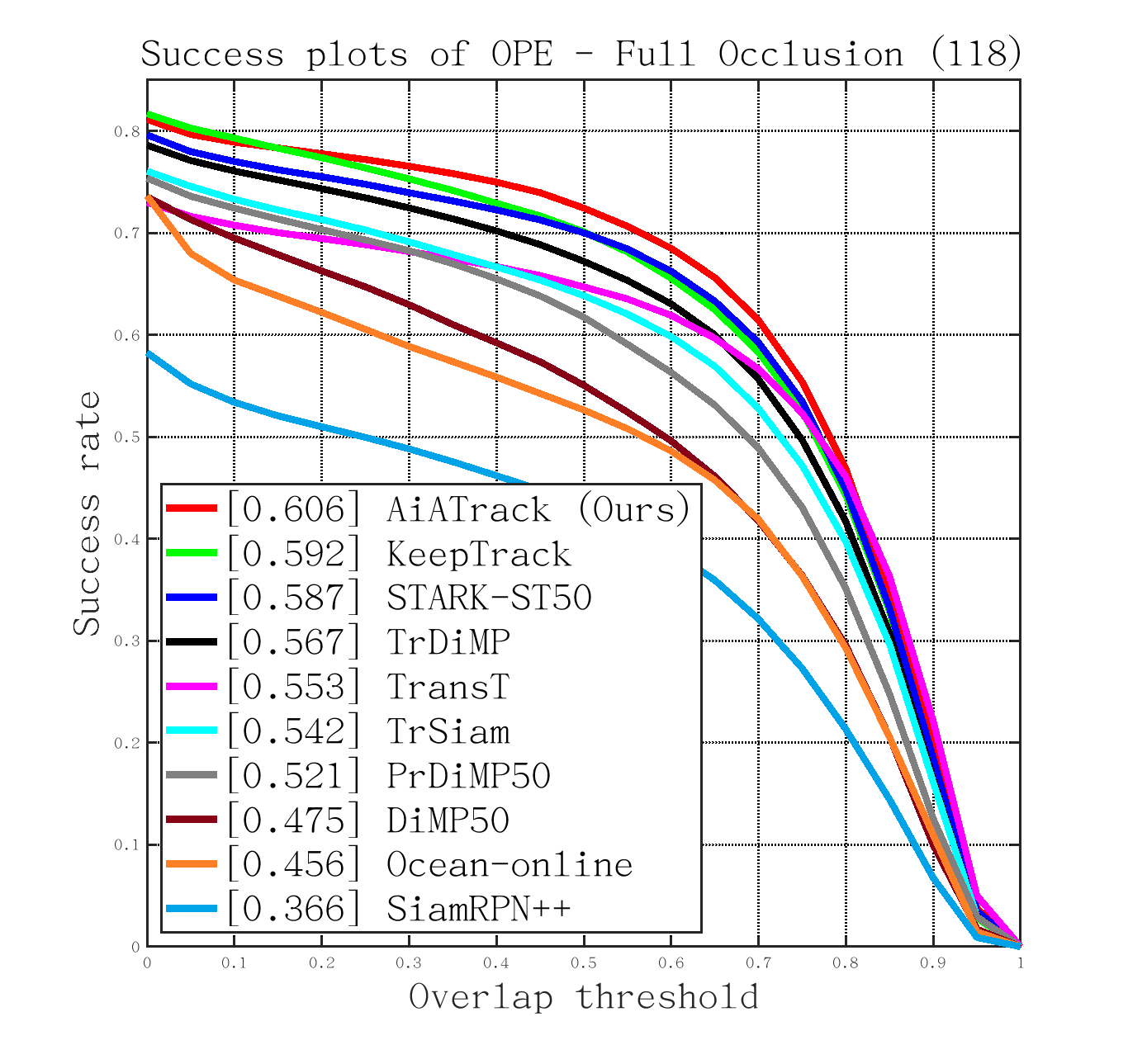}}
\enspace
\subfigure{\includegraphics[width=.28\textwidth]{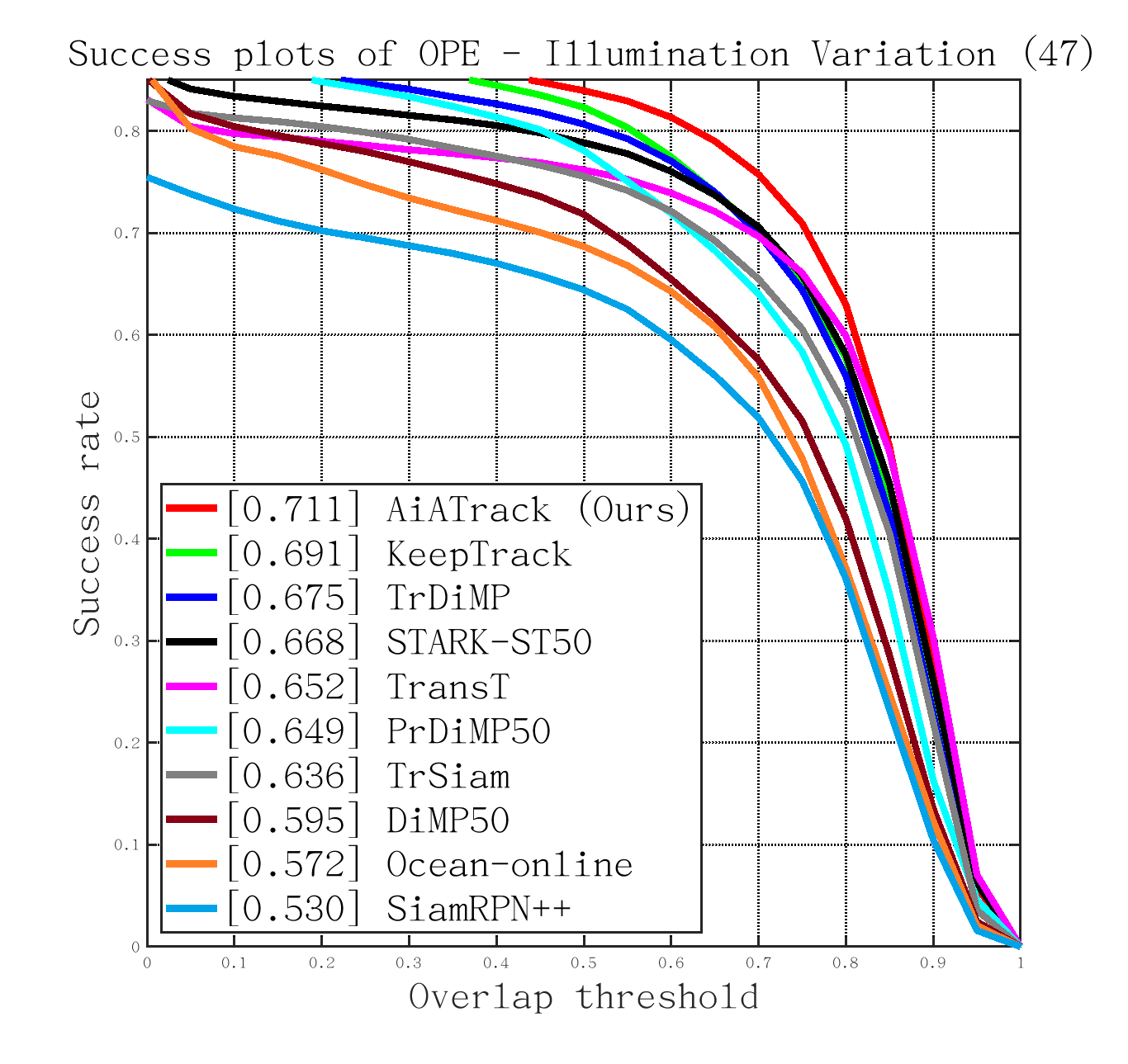}}
\enspace
\subfigure{\includegraphics[width=.28\textwidth]{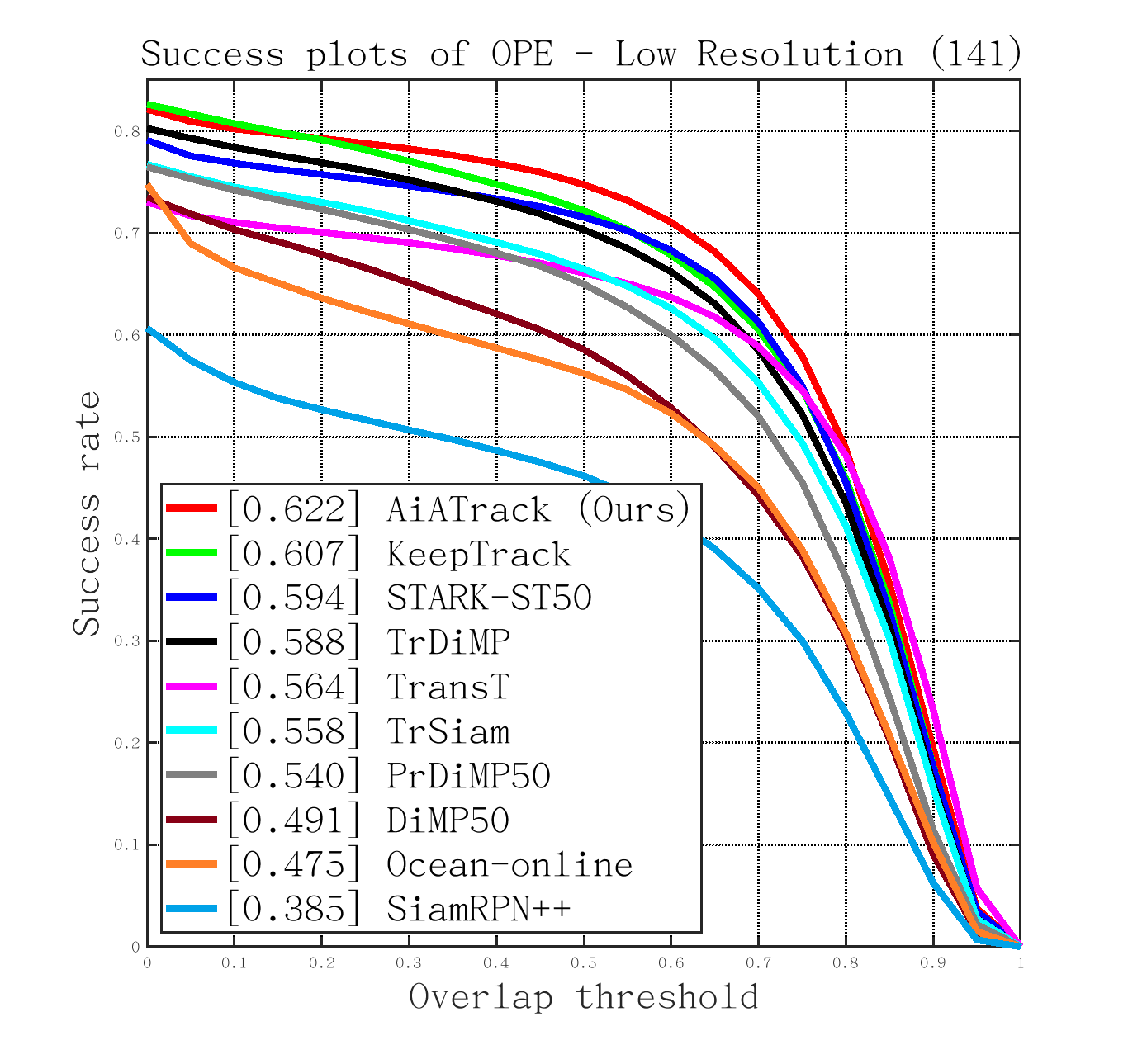}}
\enspace
\subfigure{\includegraphics[width=.28\textwidth]{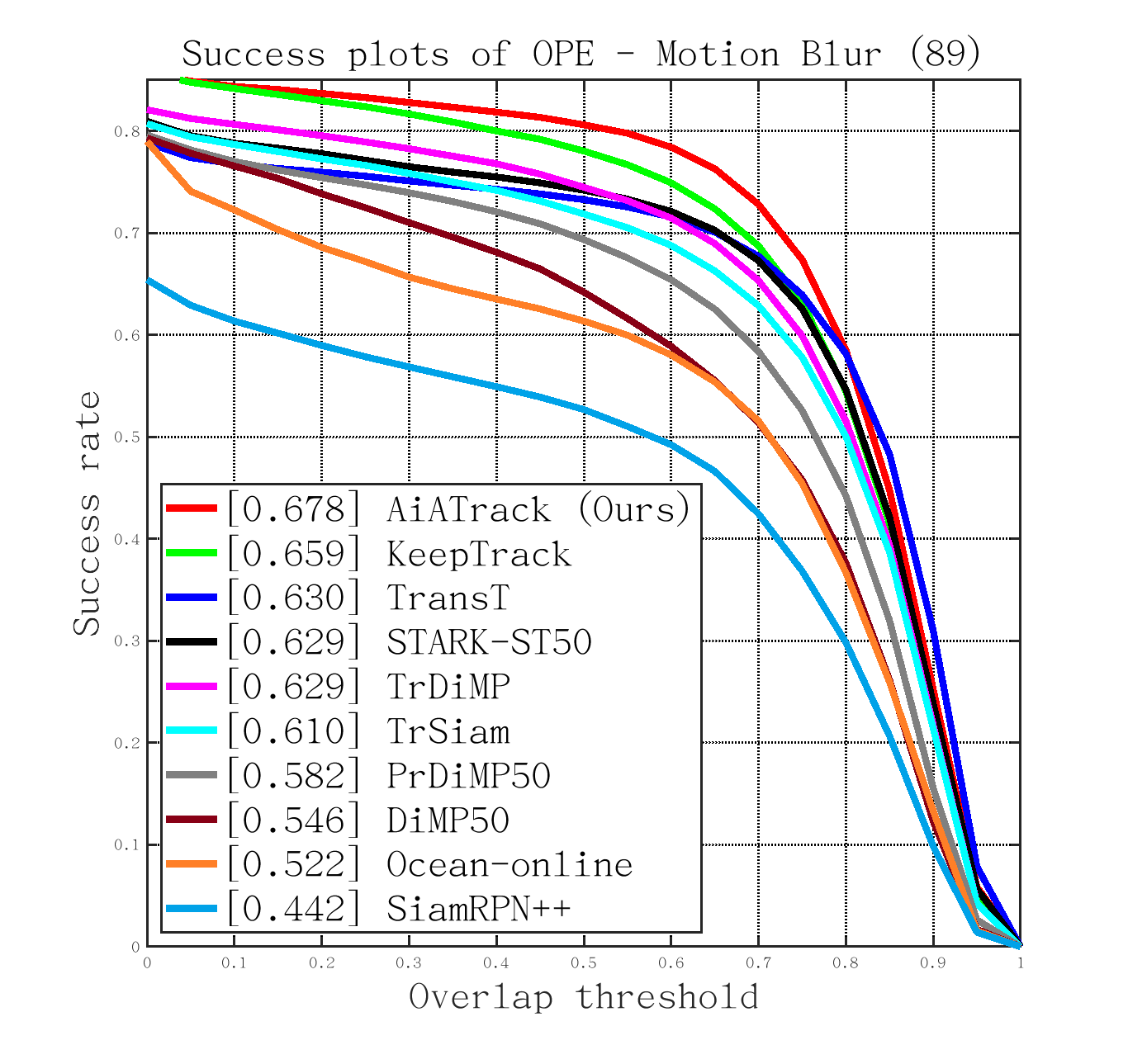}}
\enspace
\subfigure{\includegraphics[width=.28\textwidth]{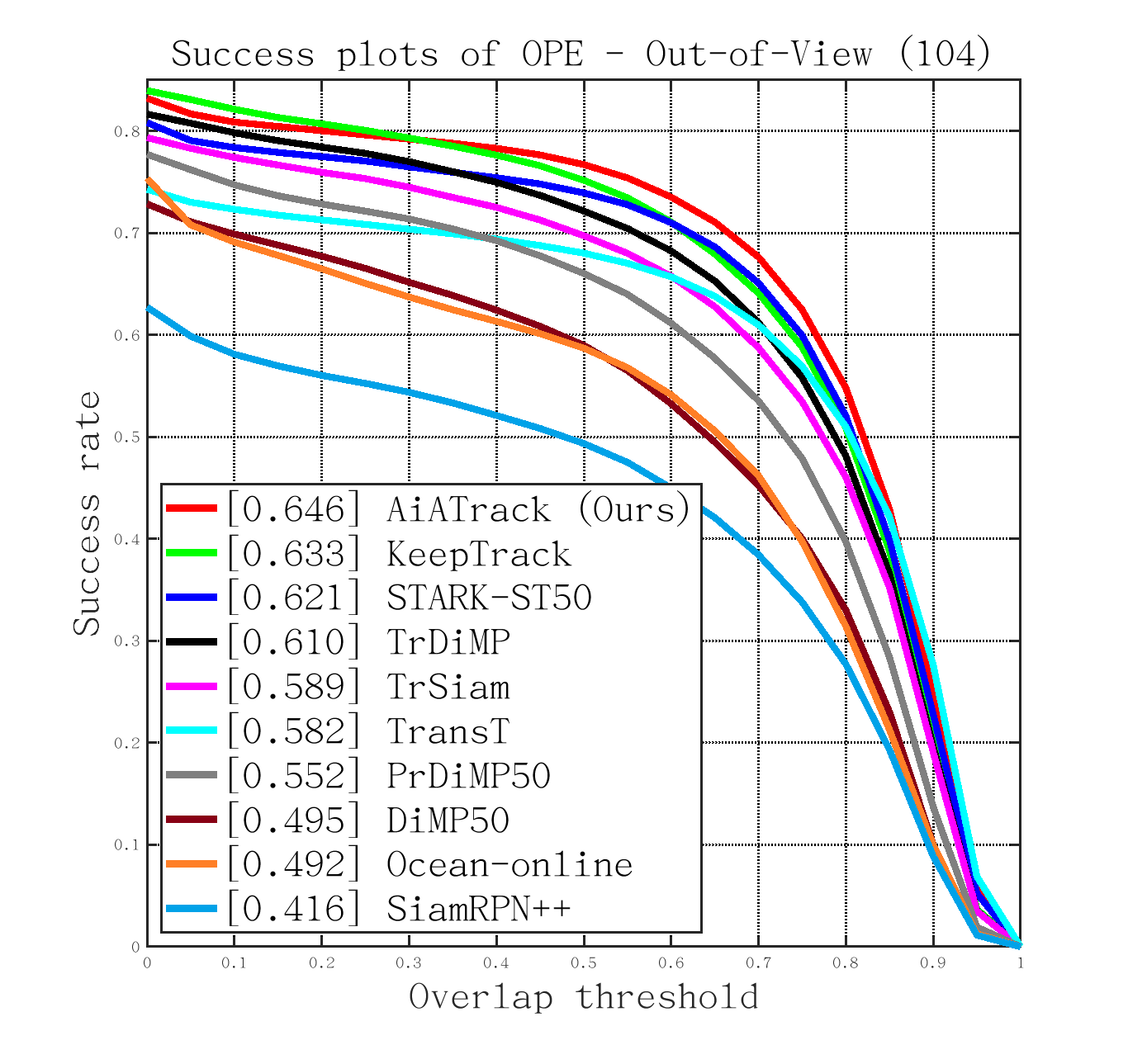}}
\enspace
\subfigure{\includegraphics[width=.28\textwidth]{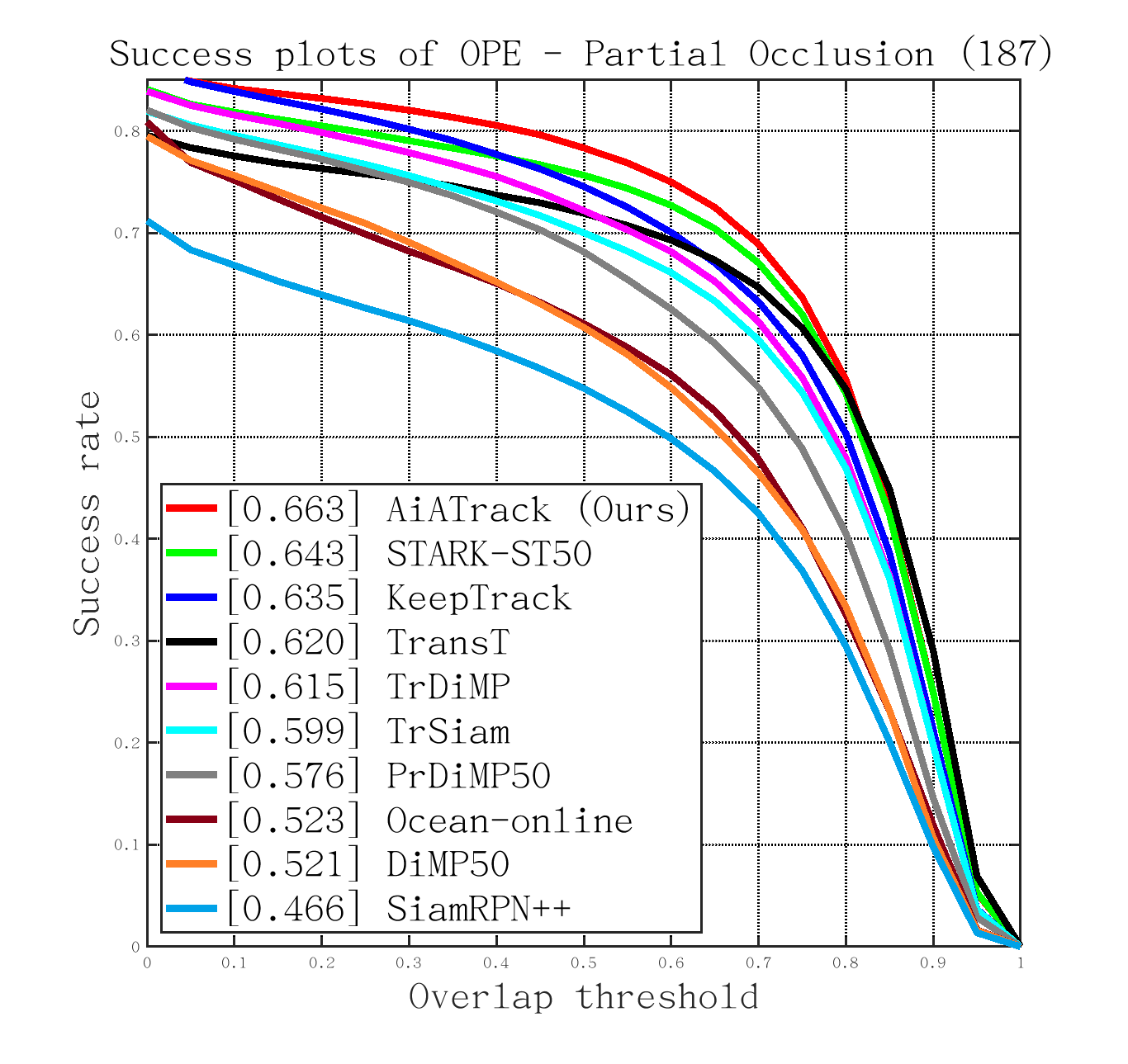}}
\enspace
\subfigure{\includegraphics[width=.28\textwidth]{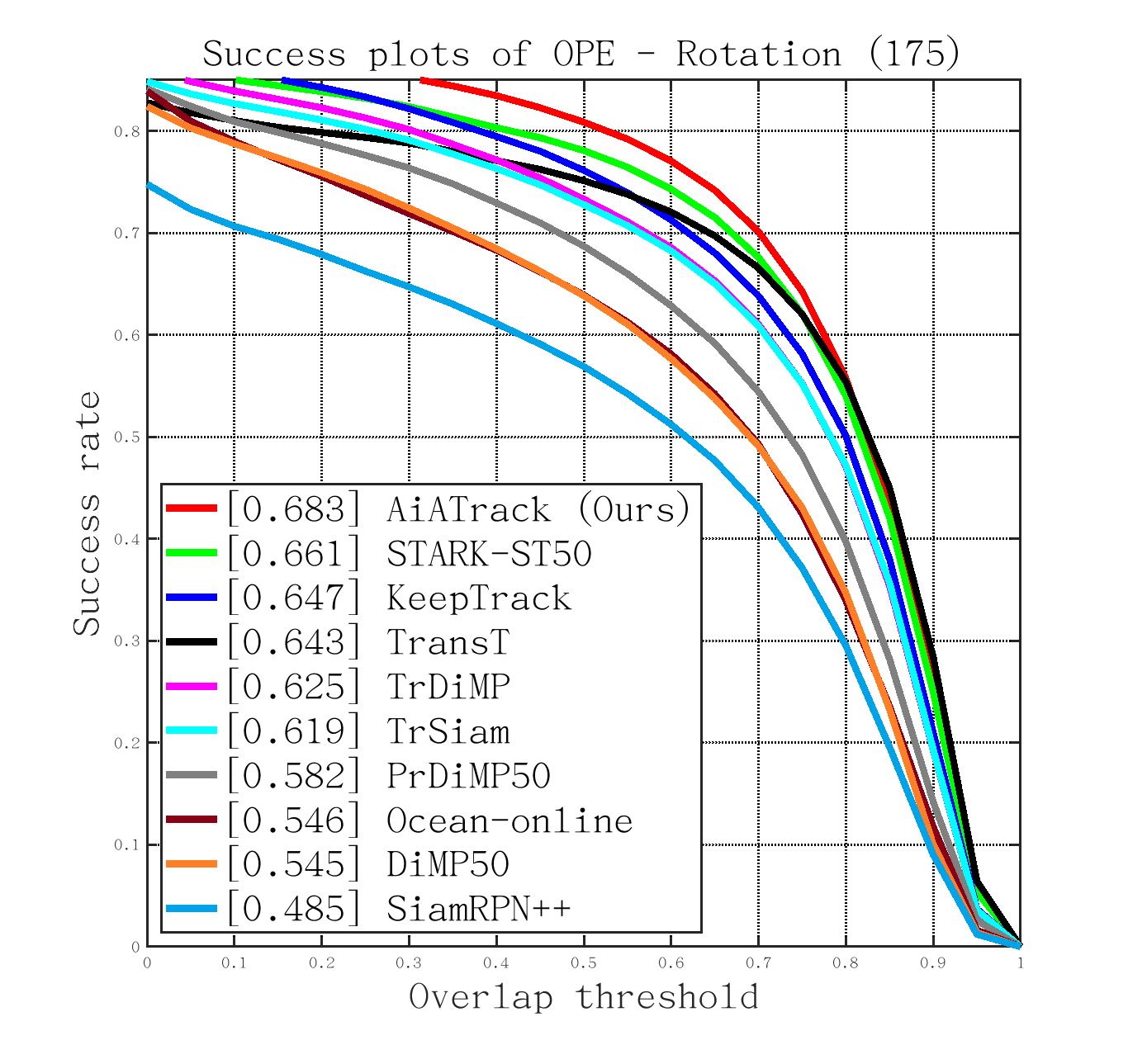}}
\enspace
\subfigure{\includegraphics[width=.28\textwidth]{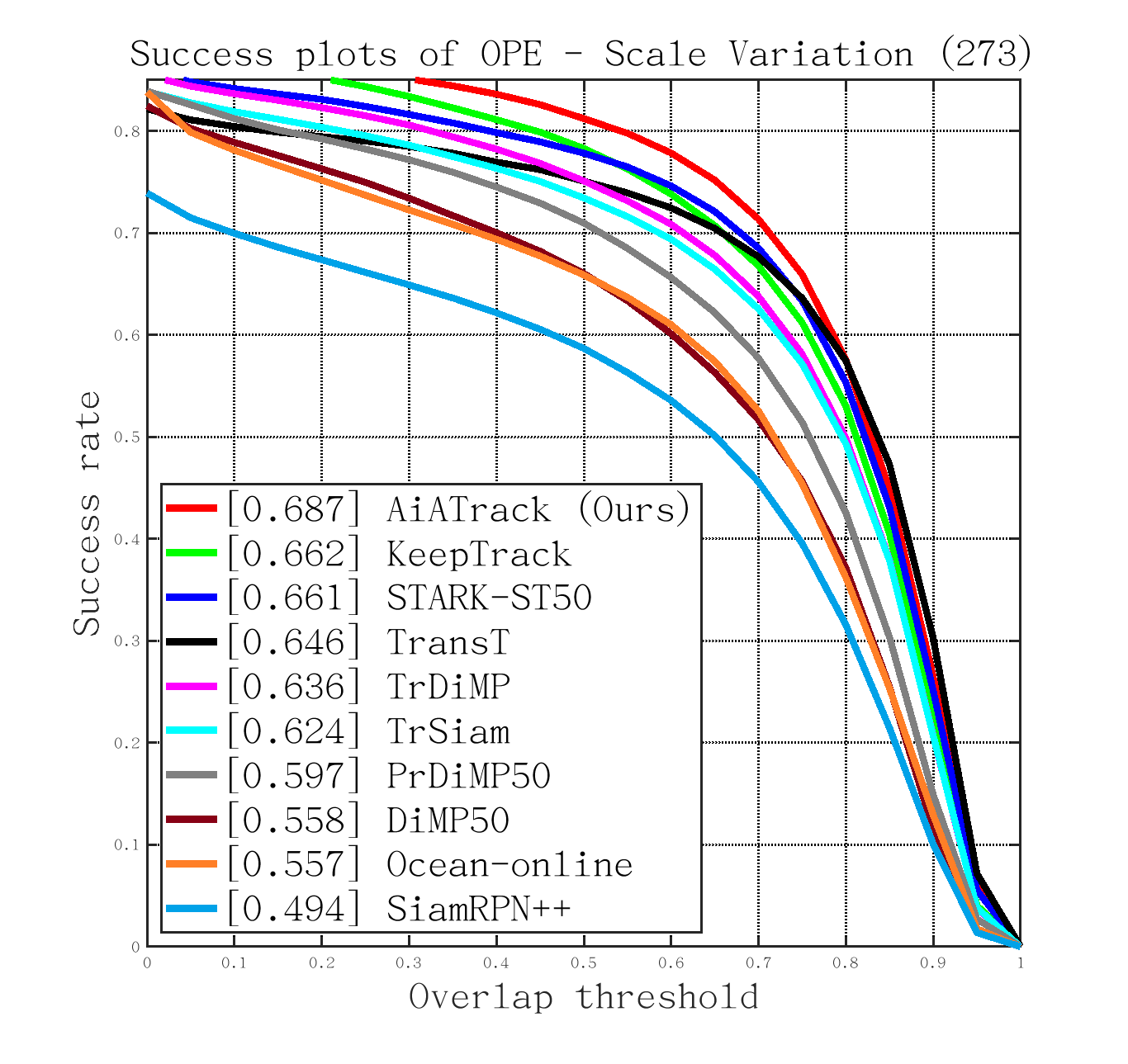}}
\enspace
\subfigure{\includegraphics[width=.28\textwidth]{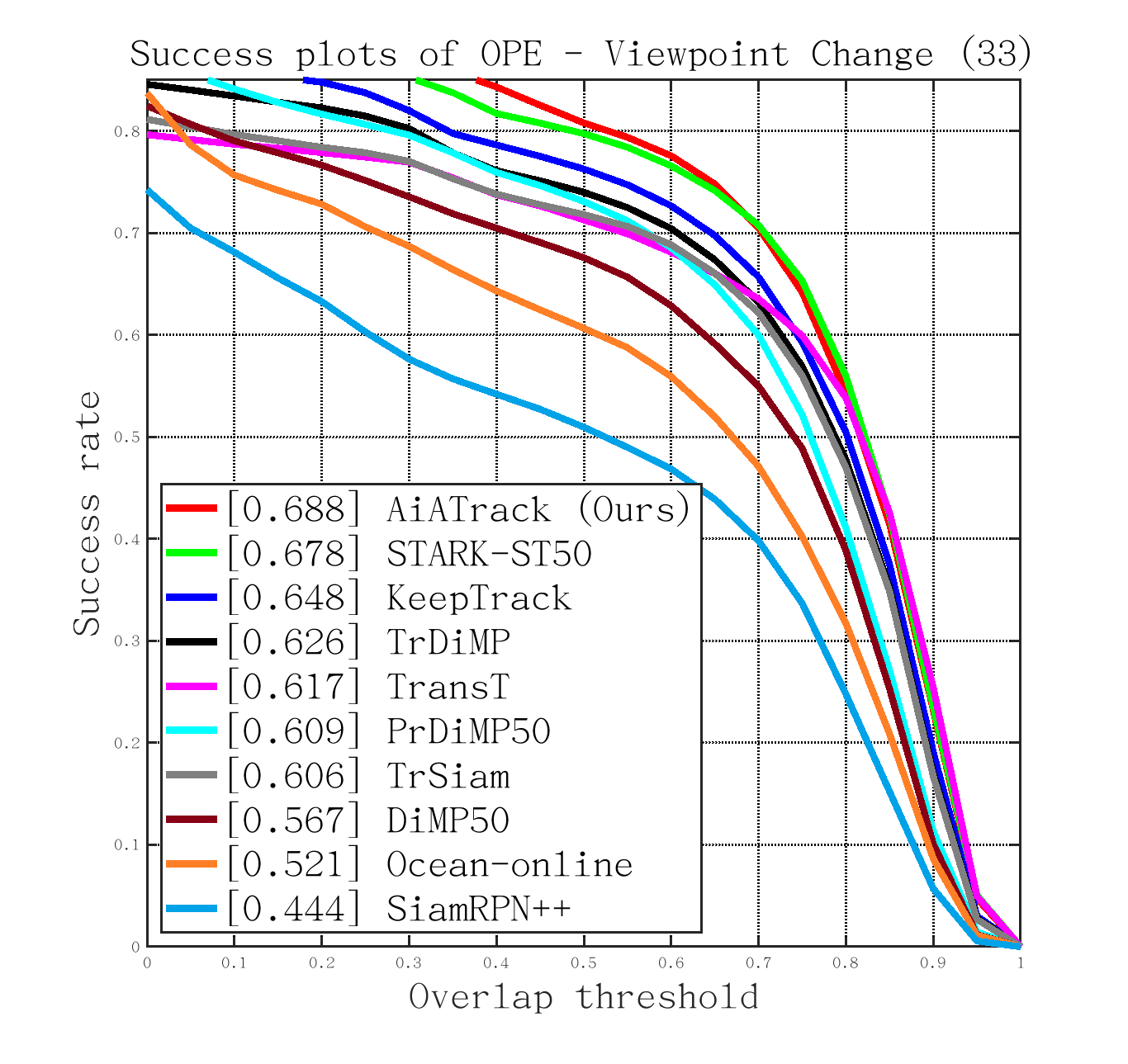}}
\caption{Attribute analysis on LaSOT. AUC scores are showed in the legend.}
\label{figure-analysis}
\end{figure}

\begin{figure}[t]
\centering
\includegraphics[width=.98\textwidth]{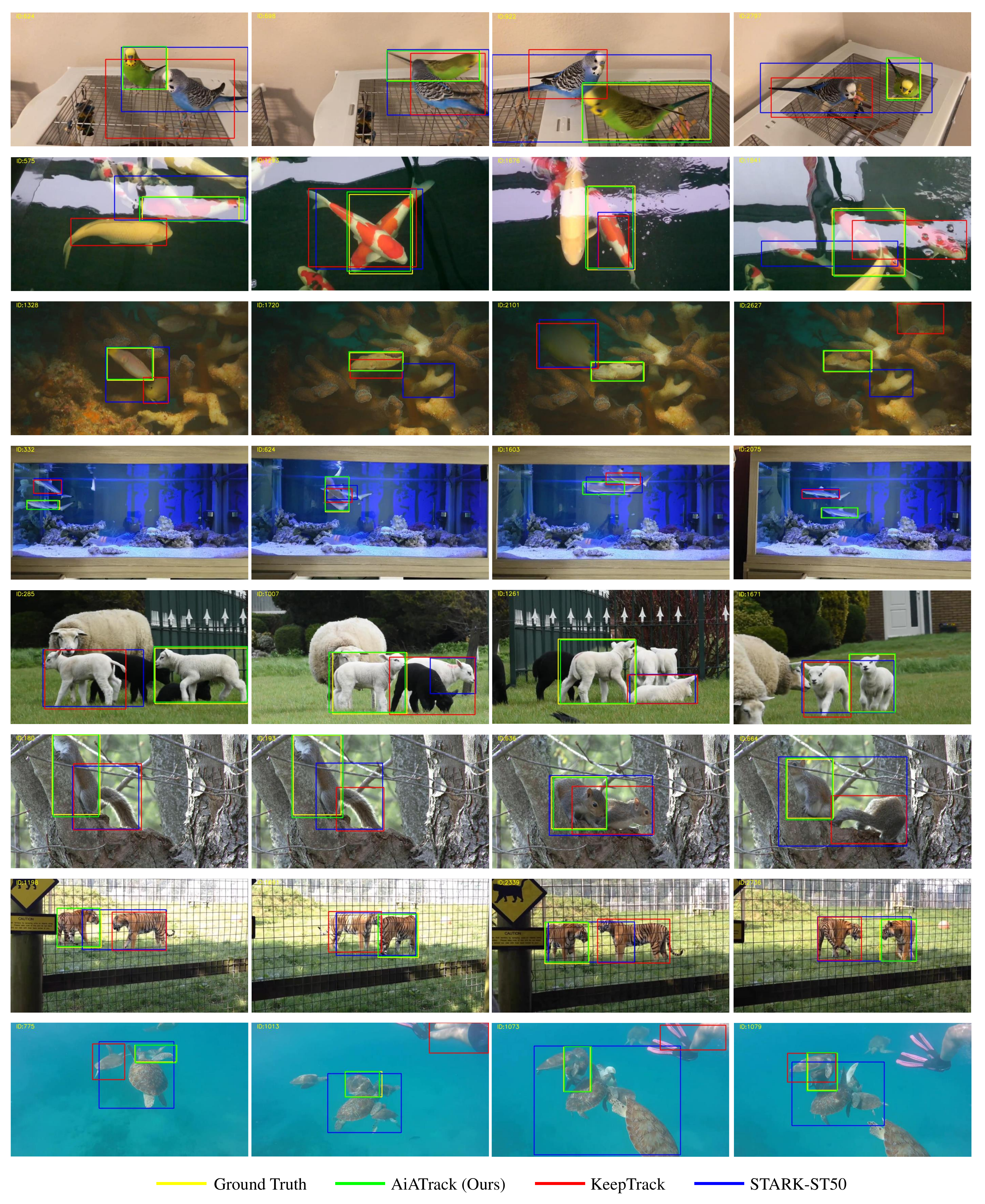}
\caption{Qualitative comparisons with two representative state-of-the-art trackers on 8 challenging sequences: \textit{bird-17}, \textit{goldfish-8}, \textit{sepia-13}, \textit{shark-2}, \textit{sheep-3}, \textit{squirrel-8}, \textit{tiger-4}, \textit{turtle-8}. Frame indexes are given on the top-left of each figure.}
\label{figure-qualitative}
\end{figure}

\subsection{Attribute Analysis}
We also provide detailed attribute analysis on LaSOT \cite{fan2019lasot}. Fig. \ref{figure-analysis} shows that our tracker has an encouraging performance in various kinds of scenarios like background clutter, camera motion, and deformation. The results suggest the great potential of the proposed method when dealing with challenging scenarios.

\subsection{Qualitative Comparisons}
To qualitatively compare our tracker with the state-of-the-art trackers, we visualize our tracking results with two recent representative trackers: KeepTrack \cite{mayer2021learning} and STARK \cite{yan2021learning}. Fig. \ref{figure-qualitative} shows the tracking outputs for these trackers on some challenging video examples.

\end{document}